\documentclass[journal]{IEEEtran}
\ifCLASSINFOpdf
\else
\fi
\usepackage{times}
\usepackage{makecell}
\usepackage[switch]{lineno}
\usepackage[table]{xcolor}

\usepackage[small]{caption}
\usepackage{graphicx}
\usepackage{amsmath}
\usepackage{booktabs}
\usepackage{multirow}
\usepackage{amsthm}
\usepackage{verbatim}
\usepackage{amsfonts,amssymb,mathrsfs}
\usepackage[subfigure]{tocloft}
\usepackage{subfigure}
\usepackage{xcolor}
\usepackage{algorithmic}
\usepackage[hidelinks]{hyperref}
\usepackage{amsmath}
\usepackage{amsfonts,amssymb,mathrsfs}
\usepackage{booktabs}
\usepackage{subfigure}

\usepackage{comment}

\begin{document}
%
\title{Towards Benchmarking and Assessing the Safety and Robustness of Autonomous Driving on Safety-critical Scenarios}


\author{Jingzheng Li, Xianglong Liu${^*}$,~\IEEEmembership{Member,~IEEE,} Shikui Wei,~\IEEEmembership{Member,~IEEE,} Zhijun Chen,\\ Bing Li, Qing Guo,~\IEEEmembership{Member,~IEEE,}  Xianqi Yang, Yanjun Pu and Jiakai Wang  
\thanks{Jingzheng Li, Xianqi Yang, Yanjun Pu, Jiakai Wang are with Zhongguancun Laboratory, Beijing 100094, China.}
\thanks{Xianglong Liu is with Zhongguancun Laboratory, Beijing 100094, China, also with the State Key Laboratory of Software Development Environment, Beihang University, Beijing 100191, China, and also with the Institute of Data Space, Hefei Comprehensive National Science Center, Hefei, China.}
\thanks{Shikui Wei is with Institute of
Information Science, Beijing Jiaotong University, Beijing, China 100044, and also with Beijing Key Laboratory of Advanced Information Science and Network Technology, Beijing 100044, China.}
\thanks{Zhijun Chen is with Beihang University, Beijing 100091, China.}
\thanks{Bing Li and Qing Guo are with A*STAR, 138632, Singapore.}
\thanks{${^*}$ Corresponding author: Xianglong Liu (e-mail:xlliu@buaa.edu.cn).}
\thanks{Manuscript received April 19, 2005; revised August 26, 2015.}}

%
%

\markboth{Journal of \LaTeX\ Class Files,~Vol.~14, No.~8, August~2015}%
{Shell \MakeLowercase{\textit{et al.}}: Bare Demo of IEEEtran.cls for IEEE Journals}
%



\maketitle

\begin{abstract}

Autonomous driving has made significant progress in both academia and industry, including performance improvements in perception task and the development of end-to-end autonomous driving systems. 
However, the safety and robustness assessment of autonomous driving has not received sufficient attention. Current evaluations of autonomous driving are typically conducted in natural driving scenarios. However, many accidents often occur in edge cases, also known as safety-critical scenarios.
These safety-critical scenarios are difficult to collect, and there is currently no clear definition of what constitutes a safety-critical scenario.
In this work, we explore the safety and robustness of autonomous driving in safety-critical scenarios. 
First, we provide a definition of safety-critical scenarios, including static traffic scenarios such as adversarial attack scenarios and natural distribution shifts, as well as dynamic traffic scenarios such as accident scenarios.
Then, we develop an autonomous driving test framework to comprehensively evaluate autonomous driving systems, encompassing not only the assessment of perception modules but also system-level evaluations.
Our work systematically constructs a safety verification process for autonomous driving, providing technical support for the industry to establish standardized test framework and reduce risks in real-world road deployment.
\end{abstract}

\begin{IEEEkeywords}
Deep learning, Autonomous driving, Safety and robustness, Adversarial attack, Distribution shift.
\end{IEEEkeywords}

%
\IEEEpeerreviewmaketitle

\section{Introduction}
%
%
%
%
\IEEEPARstart{R}{ecently}, Autonomous Driving (AD) algorithms have made remarkable progress and can accomplish driving task through a set of predefined routes~\cite{chen2024end,hu2023planning}.
Tesla and Waymo release test reports claiming that AD drives more safely than humans~\cite{boudette2021tesla}.
However, we argue that the test reports are not rigorous and are biased; because the AD systems are evaluated on common and simple driving scenarios that seen during the training phase. Actually, the accident-prone driving scenarios, aka, safety-critical scenarios, are key to ensuring that AD moves toward large-scale commercial deployment~\cite{feng2021intelligent,schutt20231001,ding2023survey}.
Currently, the safety evaluation of AD on safety-critical scenarios has not been paid enough attention. Meanwhile, the scope/definition of safety-critical scenarios is also relatively limited without a well standardized.
\begin{figure}[t]
  \centering
\includegraphics[width=0.99\linewidth,scale=1]{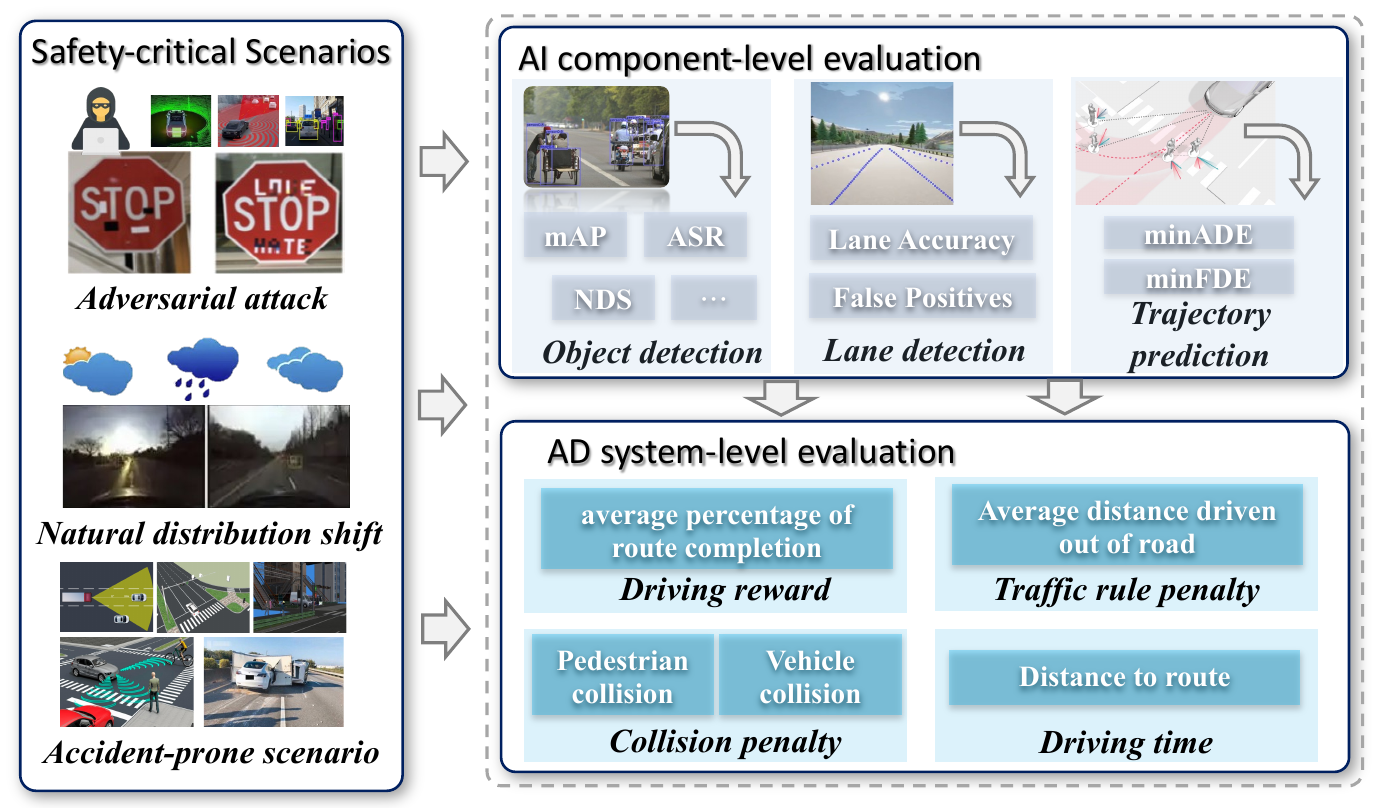}
  \caption{Safety and robustness evaluation on safety-critical scenarios including the evaluation of AI component-level such as object detection and the evaluation of the functional safety of autonomous driving system.}
  \label{framework2}
\end{figure}

Waymo reports an average of 210,000 kilometers before an accident. In real-world, the accident-prone driving scenarios are rare such that these long-tail scenarios are difficult to collect and dangerous to evaluation~\cite{feng2023dense}.
These challenges can be addressed with realistic simulation~\cite{hao2023adversarial}.
Thus, some recent work has focused on evaluating the safety and robustness of AD on the safety-critical scenarios~\cite{xu2022safebench,chen2024automated,hu2022pass}, which can be divided into two main categories: (1) one focuses on evaluating the robustness of AD' perception module on static safety-critical scenarios~\cite{wang2023does,zheng2024physical}, e.g., using adversarial attack methods to render textures of vehicles, pedestrians or traffic signs in driving scenarios leads to incorrect predictions of perceptual models such as the object detection task; (2) the other focuses on the safety evaluation of AD' planning and control modules on dynamic safety-critical scenarios~\cite{hanselmann2022king,chen2024frea}, e.g., constructing realistic and diverse accident-prone driving scenarios on natural driving scenarios based on simulation platforms.
However, these works have some limitations.

\textit{Limitations}: (1) There is a lack of standard of safety-critical scenarios. Specifically, the aforementioned adversarial attacks consider only static assets in the scenarios. Besides the security issue caused by human malicious attacks, there are also some natural distributional shift that can also lead to safety risks in the  perception modules of AD~\cite{dong2023benchmarking,zhu2023understanding,xia2024openad}, such as motion blur, severe weather and complex surroundings.
On the other hands, some works have focused on generating dynamic safety-critical scenarios that would lead to a collision by controlling the behavior of the surrounding vehicles. However, these works have only focused on a limited number of predefined scenarios such as right-turns, lane changing, etc. Thus, generating more diverse accident scenarios is also a high priority~\cite{gaomagicdrive}.
(2) There is an AI-to-system semantic gap. Considerable works develop adversarial attack methods that can effectively cause AI algorithms to make prediction errors then evaluate the robustness of perceptual module. Actually, AI component-level errors do not necessarily lead to system level effect. 
Consequently, there is an urgent need to effectively evaluate the AD system~\cite{hu2022pass}.

To address these limitations, we develop a AD safety and security testbed, SSAD, to comprehensively evaluate the behavior of AD on safety-critical scenarios.
Fig.\ref{framework2} illustrates the framework of SSAD, which mainly consists of safety-critical scenarios and safety and robustness assessment on top of it.
(1) The safety-critical scenarios in SSAD include both adversarial attack and natural distribution shift, and also include generated accident-prone scenarios. 
Among them, natural distribution shift is a type of previously overlooked safety-critical scenario, which mainly includes environment noises such as severe weather and sensor noises such as camera exposure.
SSAD can ensure a comprehensive analysis of AD systems under diverse safety-critical scenarios. 
(2) To address the AI-to-system semantic gap, the evaluation of SSAD includes not only the evaluation of perception module, i.e., the evaluation of the AI models, but also the system-level evaluation such as route completion, collision rate, etc. In addition, the evaluation of perception module includes not only  natural noise, but also a range of adversarial attack methods.
\begin{figure*}[ht]
  \centering
\includegraphics[width=0.85\linewidth,scale=1]{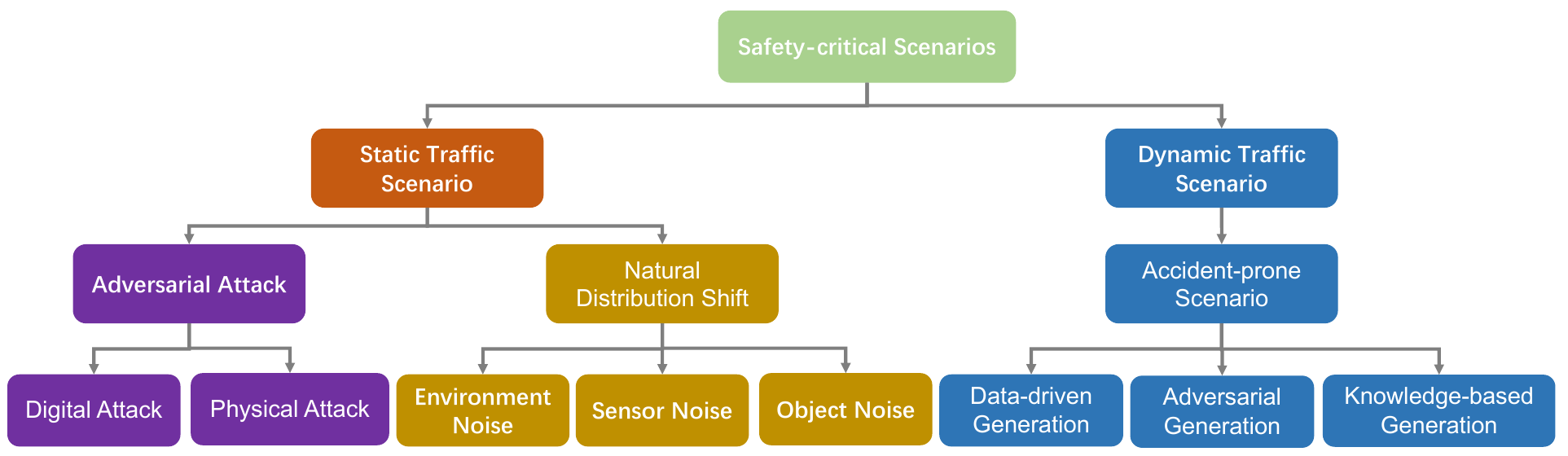}
  \caption{The categorization of safety-critical scenario includes static and dynamic traffic scenarios. static traffic scenario includes adversarial attacks and natural distributional shift. Dynamic traffic scenario is mainly accident-prone scenarios.}
  \label{motivation}
\end{figure*}

Overall, the contributions of our work can be summarized as follows:
\begin{itemize}
\item We design an evaluation framework, SSDA, which can evaluate the safety and robustness of AD on safety-critical scenarios. 
SSDA could integrate multiple perception tasks, and planning and control algorithms, and diverse safety-critical scenarios. 
\item We standardize the definition of safety-critical scenario, including not only static scenarios of adversarial attack and natural distribution shift, but also the dynamic accident-prone scenarios by controlling the behavior of the surrounding traffic participants.
\item To fill the AI-to-system semantic gap, SSDA could conduct both AI component-level evaluation and AD system-level evaluation.
\end{itemize}

\section{Related Works}
\begin{table*}[t]
\caption{Comparison of Evaluation Platforms}
\centering
{
\resizebox{\linewidth}{!}{\begin{tabular}{lccccccc}
\toprule
\multirow{2}{*}{Platform} & \multicolumn{2}{c}{Adversarial attack} & \multirow{2}{*}{Distribution Shift} & \multirow{2}{*}{Scenarios Generation Alg.}& \multirow{2}{*}{Customized Scenario}& \multirow{2}{*}{Perception}& \multirow{2}{*}{Planning and Control}\\
&Digital Attack&Physical Attack & &\\
\midrule
Scenario Runner & $\times$&$\times$&$\times$&$\times$& $\checkmark$&$\times$&$\times$\\
CommonRoad~\cite{althoff2017commonroad}& $\times$&$\times$&$\times$&$\times$& $\checkmark$&$\times$&$\times$\\
Highway-Env~\cite{highway-env}& $\times$&$\times$&$\times$&$\times$& $\checkmark$&$\times$&$\times$\\
Pylot~\cite{gog2021pylot} &$\times$&$\times$&$\times$&$\times$& $\times$&$\checkmark$&$\checkmark$\\
PASS~\cite{hu2022pass}&$\checkmark$&$\checkmark$&$\times$&$\times$& $\times$&$\checkmark$&$\times$\\
ANTI-CARLA~\cite{ramakrishna2022anti}&$\times$&$\times$&$\times$&$\checkmark$& $\times$&$\times$&$\checkmark$\\
SafeBench~\cite{xu2022safebench} &$\times$&$\checkmark$&$\times$&$\checkmark$& $\checkmark$&$\checkmark$&$\checkmark$\\
D2RL~\cite{feng2023dense}&$\times$&$\times$&$\times$&$\checkmark$& $\times$&$\times$&$\times$\\
\midrule
SSDA& \checkmark & \checkmark & \checkmark& \checkmark& \checkmark& \checkmark& \checkmark\\
\bottomrule
\end{tabular}}
}
\label{tab1}
\vspace{-0.5cm}
\end{table*}
This work focuses on the behavior of autonomous driving on long-tail distribution data, i.e., the corner cases. 
Firstly, we introduce the research of AD on the adversarial attack and natural distribution shift, then we present the related work on the dynamic safety-critical scenario generation methods, and finally, we investigate some representative platforms about safety and robustness of AD, and make a comparative analysis with our work.
\subsection{The Safety and Robustness Evaluation of Autonomous Driving}
Ensuring the safety of autonomous driving systems remains a critical challenge. 
Current research focuses on evaluating the safety of autonomous driving from various dimensions, including perception, planning, decision-making, and controller modules~\cite{gog2021pylot,ramakrishna2022anti,xu2022safebench,shen2022sok}.
These efforts can be broadly categorized into AI component-level and system-level evaluation.
For AI component-level evaluation, autonomous driving systems rely on a range of AI modules, particularly for perception tasks. 
These AI models address critical tasks such as traffic sign recognition~\cite{wang2021can}, lane detection~\cite{zhang2024towards}, and trajectory prediction~\cite{cao2022advdo}.
AI component-level evaluation often designs adversarial attack strategies or explores out-of-distribution driving data, causing AI models to deviate from expected outcomes and even resulting in accidents.
For the system-level evaluation, the core idea is to design methods that generate driving scenarios capable of inducing accidents.
These scenarios are then used to evaluate the driving performance of the ADs. The evaluation metrics typically consider three levels: (1) safety level, such as collision rate; (2) functionality level, such as average percentage of route completion; (3) etiquette level, such as average acceleration. 
PASS~\cite{hu2022pass} is a system-driving evaluation platform for AD AI security research which collect and analyze vulnerabilities of autonomous driving systems to adversarial attacks at both the AI component level and system level.
Further, SafeBench~\cite{xu2022safebench} systematically evaluates the performance of ADs over accident-prone scenarios generated by scenario generation algorithms based on naturalistic driving data. Feng et al.,~\cite{feng2023dense} propose dense deep-reinforcement-learning (D2RL) approach where the background agents learn what adversarial manoeuvre to execute so that D2RL-trained agents can accelerate the evaluation process by multiple orders of magnitude.

Tab.\ref{tab1} summarises some representative AD evaluation platforms, while highlighting the differences with SSAD.
\subsection{Adversarial Attacks and Natural Distribution shifts for Autonomous Driving}
Adversarial attacks against AD have gained significant attention in recent years.
Digital attacks modify input data by adding imperceptible perturbations in the digital space that can cause deep neural networks (DNNs) to produce incorrect outputs. Examples of digital attack methods include the Fast Gradient Sign Method (FGSM)~\cite{goodfellow2014explaining}, Projected Gradient Descent (PGD)~\cite{madry2018towards}, and Carlini-Wagner (CW)~\cite{carlini2017towards}.
Some works have also explored the effectiveness of adversarial examples in the physical world.
Physical adversarial attacks often require digital-to-physical transformations to improve robustness such as Expectation over Transformation (EoT), non-printability score (NPS) loss, and total variant (TV) loss.
Physical adversarial attacks in the real world are  observed by~\cite{wu2020physical} where they found
that images with adversarial permutation printed out on paper remain effective.
There have been
many physical attack studies that attack deep learning algorithms.
For example, Athalye et al.,~\cite{athalye2018synthesizing} propose EoT and construct physical attacks by with 3D-printed objects. Evtimov et al.,~\cite{eykholt2018robust} use black and white stickers to attack stop signs. Tu et al.,~\cite{eykholt2018robust} craft adversarial mesh placed on top of a vehicle to bypass a LiDAR-based detection. 
Cao et al.,~\cite{cao2021invisible} reveal the possibility of crashing multi-sensor fusion based models by attacking all fusion sources simultaneously.
Sato et al.,~\cite{sato2024intriguing} use the diffusion model to generate images with different patterns such as traffic signs, which have natural attack capability against DNN models, spoof the perception module of Tesla Model 3.
It has been shown that DNN-based object detections can be vulnerable to adversarial examples~\cite{xie2023adversarial,zhu2023understanding} across LiDAR-based, camera-based, or fusion models.


Natural distribution shifts, where the test data differs from the training data due to variations in environmental factors, pose significant challenges to the robustness and safety of autonomous driving systems.
Breitenstein et al.,~\cite{breitenstein2020systematization} propose a structured classification of distribution shift in ADs, which spans multiple levels, ranging from pixel-level issues like overexposure and defective pixels, to domain-level shifts caused by varying weather conditions, and up to scenario-level cases that involve temporal context and potential for collision. 
Heidecker et al.,~\cite{heidecker2021application} review existing definitions of distribution shift from outlier, novelty, anomaly, and out-of-distribution detection literature, and provide a detailed categorization of corner cases into different layers (sensor, content, temporal) and levels (hardware, physical, domain, object, scene, scenario), highlighting specific examples for each sensor type. 
Dong et al.,~\cite{dong2023benchmarking} systematically evaluate the robustness of 3D object detections to common corruptions encountered in real-world scenarios. The authors design 27 types of corruptions covering weather conditions, sensor noises, motion distortions, object deformations, and sensor misalignment, and apply them to available datasets to create three corruption robustness benchmarks. 
Further, Li et al.,~\cite{li2024r} provide a systematic evaluation framework for assessing the robustness of perception module against diverse types of perturbations. 
Similarly, Li et al.,~\cite{li2022coda} propose CODA dataset by providing a realistic and diverse collection of corner cases for evaluating object detections in autonomous driving.
We summarize previous definitions of distribution shift in AD and present a categorization: ‌environmental noise, sensor noise and ‌object noise.


\subsection{Accident-prone Driving Scenario
Generation}
Unlike the aforementioned works, which aim to generate realistic and natural scenarios~\cite{li2023scenarionet,wei2024editable}, we focus on the long-tail distribution,
consisting of the safety-critical scenarios, to provide efficient
evaluations of the AV safety. 
We follow the previous work~\cite{ding2023survey} to divide the scenario generation methods into 3 categories as follows.
(1) The data-driven scenario generation approaches~\cite{knies2020data,scanlon2021waymo} leverage real-world driving datasets to extract or synthesize rare accident scenarios by using generative models.
However, the collected data is highly unbalanced regarding safe and risky scenarios, which makes it challenging to train generative models to generate safety-critical scenarios.
(2) The most dominant approach is currently adversarial-based methods, which mainly builds a policy model to control dynamic objects or modify the environment state to interfere with the inputs of the ego vehicle and thus influence the decision-making process. Therefore, the adversarial-based scenario generation algorithms can be divided into driving policy-based and transition function-based adversarial generation.
For the driving policy-based adversarial generation, it is usually formulated as a Reinforcement Learning (RL) problem, where the ego vehicle belongs to the environment and the generator is the agent we can control.
Feng et al.,~\cite{feng2021intelligent} and Sun et al.,~\cite{sun2021corner} use deep
Q-network to generate discrete adversarial traffic scenarios.
Chen et al.,~\cite{chen2021adversarial} uses
Deep Deterministic Policy Gradient (DDPG) to generate adversarial policy to control surrounding agents to generate lane-changing scenarios.
Wachi et al.,~\cite{wachi2019failure} use multi-agent DDPG to control two surrounding vehicles to attack the ego vehicle.
For the transition function-based methods, some works~\cite{wang2021advsim,hanselmann2022king,zhang2023cat} consider that the environment state can be directly obtained through the Kinematics in simulator, so that the Kinematics can be maliciously modified to generate complex environment states by using gradient updating.
Some works~\cite{xu2023diffscene,huang2024cadre} take into account the future trajectories of the surrounding traffic participants considered in the environment state and therefore design some adversarial attack methods to generate malicious trajectory predictions thereby influencing the decision making of the ego vehicle.
(3) Knowledge-based methods~\cite{cai2020summit} rely on ‌explicit domain knowledge‌, e.g., traffic rules, accident statistics, or expert heuristics, to design or validate safety-critical scenarios for ADs. These methods prioritize interpretability, compliance with regulations, and alignment with real-world risk patterns. This paper mainly focuses on the adversarial-based scenario generation methods due to its good adaptability in various scenarios and effective construction of accident scenarios.
In experiments, we use driving policy-based adversarial generation method to generate safety-critical scenarios for evaluating ADs, while also utilizing knowledge-based  scenarios ``pre-crash safety-critical scenarios'' for functional testing.

\section{Safety-Critical Driving Scenario
Generation}
\label{sec3}
As shown in Fig.\ref{motivation}, we reorganize and define safety-critical scenarios, including static scenario generation, i.e., adversarial attack and natural distribution shift scenarios, and dynamic scenario generation, i.e., accident-prone scenarios.
\subsection{Static Scenarios Generation}

\textit{1) Adversarial Attack-based Scenarios}\\
\textbf{Digital Attack:}
Digital adversarial attacks~\cite{goodfellow2014explaining,carlini2017towards} refer to modify input data by elaborately crafting perturbations in digital space that is imperceptible to human observers but can cause the DNN to produce incorrect output. 
Specifically, for a well-trained
DNN $f$, the adversary generate a adversarial example $\boldsymbol{x}^{adv}$
for a given image $\boldsymbol{x}$, formalized as follows
\begin{equation}
\left\{\begin{array}{l}
\boldsymbol{x}^{adv}=\boldsymbol{x}+\boldsymbol{\delta} \\
y^{\prime}=f\left( \boldsymbol{x}^{adv};\boldsymbol{\delta}\right)
\end{array} \text { s.t. } y^{\prime} \neq y,\right.
\end{equation}
where $y^{\prime}$ denote the output of DNN $f$ on adversarial example $\boldsymbol{x}^{adv}$ and $y$ the true label.
Digital attack aims to find the best perturbation $\boldsymbol{\delta}$ so that the adversarial examples $\boldsymbol{x}^{adv}$ are misclassified. 
Mathematically, the loss function $L(y^{\prime},y)$ is maximized with respect to $\boldsymbol{\delta}$, 
\begin{equation}
\begin{aligned}
& \max _{\boldsymbol{\delta}} L(y^{\prime}, y ; \boldsymbol{\theta}) \\
& \text { s.t. }\|\boldsymbol{\delta}\|_p<\epsilon, \boldsymbol{x}^{adv}=\boldsymbol{x}+\delta \\
& x^{adv} \in[0,1]^d
\end{aligned}
\end{equation}
where the $\left\|\cdot\right\|_p$ denote the $L_P$-norm, 
$\epsilon$ controls the maximum allowable magnitude of $\boldsymbol{\delta}$ under the constrain of $L_P$-norm.

After determining the victim DNN models, the adversary has to
model the problem in terms of the available information of
the target victim model (adversary's knowledge) and decide their requirements (adversarial specificity). 
The former can be further
divided into \textit{white-box attacks}~\cite{madry2017towards} and \textit{black-box attacks}~\cite{moosavi2017universal}. 
The
latter can be divided into \textit{targeted attack} and \textit{non-targeted attack}.
White-box attacks assume that the adversary can access the full knowledge about the target victim DNN models.
Black-box attacks hypothesize that the
adversary has no knowledge about the target victim DNN models but can query the target model.
Non-targeted attacks are designed to fool the target victim DNN model to produce the incorrect prediction except the true label.
Targeted attacks are tailored to mislead
the target victim model to output the specific class assigned
by the attacker.

\begin{figure}[t]
\centering     
\subfigure[Patch-based attack: the patch generated by the adversarial attack algorithm is stuck to the target vehicle.]{\label{patch1}\includegraphics[width=0.98\linewidth]{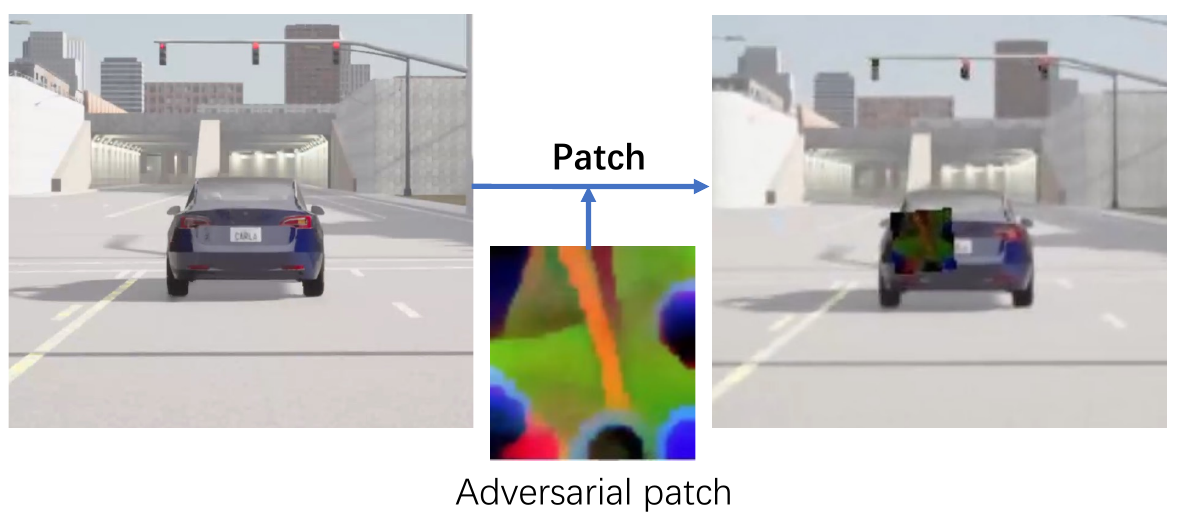}}
\subfigure[Camouflage-based attack: 
the adversarial texture generated by the adversarial attack algorithm on the mesh is rendered to the target vehicle.]{\label{patch2}\includegraphics[width=0.98\linewidth]{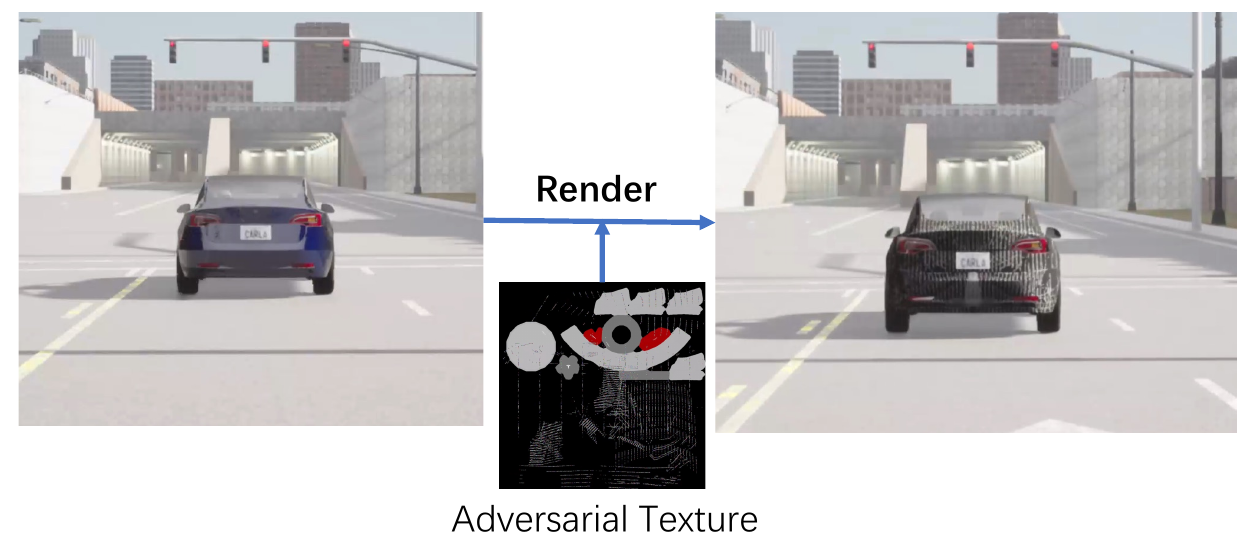}}
\caption{The process of physical attack methods on target vehicles.}
\label{patch}
\end{figure}
\textbf{Physical Attack:}
Digital adversarial examples with $L_P$ perturbations are infeasible for physical attack due to pixel-wise modifications.
In practice, the adversarial attacks~\cite{brown2017adversarial} often need to be deployed in the real world. 
However, adversarial attacks are applied to 3D objects in the physical world to significantly degrade the detection score of that object in a variety of transformations, such
as object materials, camera poses, lighting conditions, and
background interactions.
Thus, the realization of physical adversarial attacks needs to consider digital-to-physical transformation to improve the robustness of adversarial attacks.
These digital-to-physical transformation techniques mainly include expectation over transformation (EOT)~\cite{athalye2018synthesizing}, non-printability score (NPS)~\cite{sharif2016accessorize}, total variant (TV) loss~\cite{mahendran2015understanding}, etc.


A necessary common denominator for all physical attack methods is the need for a physical entity (target object) to carry the specially designed perturbations.
According to whether it requires the adversary to approach and modify the target object or not at the deployment stage of physical attacks, the physical attacks deployment can be categorized into invasive attacks
and non-invasive attacks.

\textit{Invasive Attacks}: Invasive attacks require the attacker
to approach the target object and modify its appearance with
adversarial perturbation, which can be further grouped into
\textbf{patch-based attack}~\cite{shrestha2023towards} and \textbf{camouflage-based attack}~\cite{wang2021dual,zhang2023boosting} according to the form
of perturbation. Patch-based attacks engender a universal adversarial image patch, which is stuck on the target object's surface to mislead the DNNs. Thus, patch-based
attacks are more in 2D image space. In performing patch-based physical attacks, the adversary needs to print out the
patch image with a printer and then stick/hang it on the surface
of the target object, covering its original appearance. 
Fig.\ref{patch}(a) intuitively shows the process of patch-based attack on the target vehicle.
Camouflage-based attacks generate the
adversarial texture, which is wrapped/painted over the 3D
model. Thus, camouflage-based attacks mainly rely on the physical
renderer to render the adversarial texture toward the 3D object
iteratively. In performing physical attacks, the attacker first
makes the adversarial texture physically, then wraps them over
the target object’s surface, and the original texture is covered.
Fig.\ref{patch}(b) shows the process of camouflage-based attack in the test phase.

\textit{Non-invasive Attacks}: Non-invasive attacks~\cite{zhong2022shadows,gnanasambandam2021optical} do not necessitate the attacker to physically approach and modify the target object. Instead, the attacker leverages the lighting source
to perform physical adversarial attacks, which can be done
away from the target object.
By exploiting light-based vectors‌, such operations can be executed from a distance using optical interference mechanisms‌, such as the projector~\cite{huang2022spaa,lovisotto2021slap}, the laser emitter~\cite{duan2021adversarial}, and the flashlight~\cite{gnanasambandam2021optical}. Recently, natural phenomena
can be utilized to perform physical adversarial attacks, such as
shadow~\cite{zhong2022shadows}.

\begin{figure}[t]
\centering     
\subfigure[Environment noise: left is foggy, right is rainy.]{\label{patch1}\includegraphics[width=0.98\linewidth]{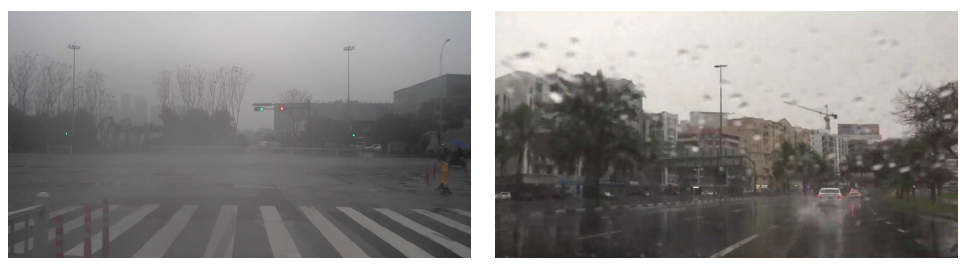}}
\subfigure[Sensor noise: left is motion blur, right is exposure.]{\label{patch2}\includegraphics[width=0.98\linewidth]{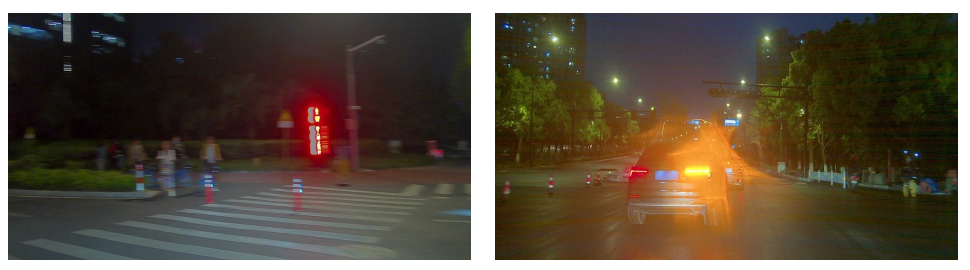}}
\subfigure[Object noise: left is novel object, right is anomalous object.]{\label{patch3}\includegraphics[width=0.98\linewidth]{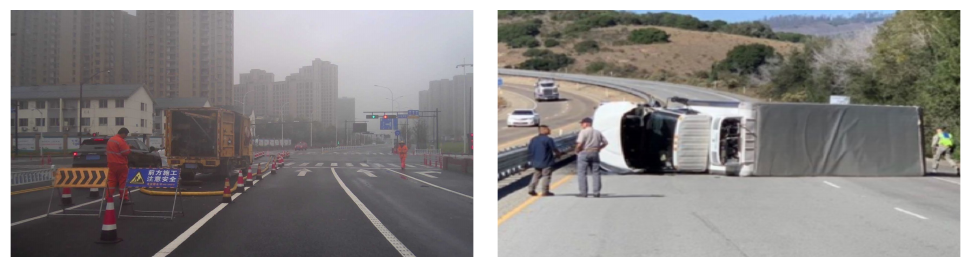}}
\caption{Some real examples of natural distribution shift. Images are from the CODA dataset~\cite{li2022coda}.}
\label{noise}
\end{figure}

\textit{2) Natural Distribution Shift-based Scenarios}\\
In addition to adversarial attacks, which are manually malicious scenarios, what is more likely to happen during real-world testing of AD is driving scenarios with natural distribution shifts. 
In other words, ADs will encounter driving scenarios during testing that are rarely seen during training phase, such as server environments and weather, different scene styles and anomaly target objects~\cite{li2022coda}. 
As depicted in Fig.\ref{motivation}, in the context of driving scenarios, natural distribution shift can be categorized into three classes according to their
origins: environment noise, sensor noise and object noise. 

\textbf{Environment Noise}: The environment noise refers to the unintended disturbance from the surrounding environment during the signal capture process. 
Severe weather and complex surrounding environment are common noises of this type.
We consider four common types of environmental noise: snow, rain, fog, and sunlight.

\textbf{Sensor Noise}: Sensor noise refers to the unwanted disturbances in the output of sensors that arise from inherent limitations or imperfections. Common sensor noises are uniform noise, gaussian noise and impulse noise.

\textbf{Object Noise}: Object noise refers to the diversity of shapes and states, viewing angles of the object itself, making it difficult for the perception algorithm to correctly recognize it.
We present 4 main object noises: scale, shear, rotation and motion blur.

In Tab.\ref{tab2}, we summarize aforementioned natural distribution shift which can be deployed in both open-loop and -loop AD test environments.
In Fig.\ref{noise}, we provide some real examples of natural distribution shift, respectively.
In reality, collecting such corner cases of distribution shift are difficult and labeling them are expensive. 
%
Thus, we provide a toolbox to convert generic autonomous driving dataset into robustness benchmark with distribution shift. which is described in Sec.IV.B.

\begin{table}[t]
\small
\caption{Different types of natural distributional shift.}
\centering
{\resizebox{0.8\linewidth}{!}{
\begin{tabular}{c|c|c}
\toprule
Environment Noise&Sensor Noise&Object Noise\\
\midrule
Snow&Uniform Noise&Scale\\
Rain&Gaussian Noise&Shear\\
Fog&Impulse Noise&Rotation\\
Sunlight&-&Motion Blur\\
\bottomrule
\end{tabular}
}}
\label{tab2}
\end{table}
\subsection{Dynamic Scenarios Generation}
Unlike static scenario generation where the safety and robustness of the AD is evaluated mainly by affected the perception module, dynamic scenario generation directly affects the planning and control modules by controlling the behavior of surrounding vehicles, which directly leads to driving accidents.
The accident-prone driving scenario generation methods can be divided into three main categories: data-driven approaches, adversarial-based methods and knowledge-based methods~\cite{ding2023survey}.
Specifically, the adversarial-based generation methods are currently the most effective method for dynamic scenario generation, which consists of two components, one is the generator, i.e, adversarial agent, and the other is the victim model, i.e., the ego agent.
The optimization problem can be considered as training a specific behavior policy in the
simulation environment, which will let adversarial agent aggressively interact with the ego agent.
In the
following, we formally present the task setting of the adversarial-based generation methods, and then detail the parameterization and objective function
used for scenario generation.

In our scope, the driving
task can be formulated as Markov Decision Process (MDP)~\cite{sutton1998reinforcement} in the form of $(S, A, R, f)$. $S$ and
$A$ denote the state and action spaces, respectively. 
$S$ includes maps sensor readings such as camera
images or LiDAR point cloud, high-level navigation commands and vehicle states. $A$ consists of
low-level control commands like steering, throttle and brake.
$R$ denotes the reward function of driving policy.
$f$ is the transition
function which outputs the state of the environment perceived by the agent, i.e., describes the dynamics of the traffic scenario.
The goal of AD is to maximize the expected reward
$\mathcal{J}(\pi, f; \boldsymbol{\theta})=\mathbb{E}_{\tau \sim \pi}\left[\sum_{t=0}^T R\left(s_t, a_t; \boldsymbol{\theta}\right)\right]$, where the driving policy $\pi$ receives the state of agent within the time horizon $T$,
$\tau \sim \pi$ is short handed for $a_t \sim \pi(\cdot|s_t; \boldsymbol{\theta})$ and $s_{t+1} \sim f(\cdot|s_t, a_t; \boldsymbol{\theta})$.
Different from the standard training of ego agent, the adversarial generation aims to minimize the reward by optimizing the environment parameter $\boldsymbol{\theta}$ for the driving policy of adversarial agent or the transition function, formalized as follows
\begin{equation}
\boldsymbol{\boldsymbol{\theta}}^*=\underset{\boldsymbol{\boldsymbol{\theta}}}{\operatorname{argmin}} \mathcal{J}(\pi, f; \boldsymbol{\theta})
\label{eq3}
\end{equation}

It can be seen that the objective function contains the transition function and driving policy parameterized by $\boldsymbol{\theta}$; the object can be minimized by optimizing the parameter.
Thus, the adversarial generation can be categorized into two ways, i.e., driving policy-based adversarial generation and   transition function-based adversarial generation.

\textit{1) Driving Policy-based Adversarial Generation}\\
Reinforcement learning is a powerful tool to solve MDP
and find optimal or suboptimal policies for learning agent.
To achieve driving policy-based adversarial generation, the first is to design the reinforcement learning algorithm to optimize the objective function; the second is to design the cost function that enforces the adversarial agent could collide with the ego agent and design some regularization to ensure that the generated safty-critical scenarios is reasonable~\cite{wachi2019failure}.

We take the L2C proposed by Ding et al.~\cite{ding2020learning} as an example, and introduce the use of reinforcement learning algorithm REINFORCE to solve the optimization problem as Eq.\ref{eq3}.
By optimizing the parameters of driving policy while freezing the parameters of the transition function, the gradient for updating driving policy is
\begin{equation}
\begin{aligned}
\nabla_{\boldsymbol{\theta}} \mathcal{J}(\pi, f; \boldsymbol{\theta}) & =E_{a \sim \pi_\theta}\left[\nabla_\theta \log \left(\pi(a; \theta)\right)\right] \mathcal{R}(s, a; \theta) \\
& \approx \frac{1}{T} \sum_i^T \nabla_\theta \log \left(\pi(a_t;\theta)\right) \mathcal{R}\left(s_t, a_t; \theta\right)
\end{aligned}
\end{equation}

What follows is the design of a cost function to encourage collisions, which mainly consists of the distance between the ego agent and the adversarial agent, as well as constraining the adversarial agent to obey traffic laws to prevent deviations from drivable areas. 
For instance, Hanselmann et al.~\cite{hanselmann2022king} design the reward function as follows
\begin{equation}
\begin{aligned}
\mathcal{R}\left(s, a\right)=\phi_{\text {col }}^{\text {ego }}(s, a)+\lambda \phi_{\text {col }}^{\text {adv }}(s, a)+\gamma \phi_{\text{dev}}^{\text {adv}}(s, a)
\end{aligned}
\end{equation}
in which $\phi_{\text {col }}^{\text {ego }}(s, a)$ denote the distance between the ago agent and adversarial agent, $\phi_{\text {col }}^{\text {adv }}(s, a)$ the safe distance between adversarial agents to improve the physical plausibility of the scenarios, $\phi_{\text {dev }}^{\text {adv }}(s, a)$ the Gaussian potential to prevent the adversarial agents from deviating from drivable areas.
Alternatively, Chen et al.,~\cite{chen2024frea} replace the collision reward with the goal-based adversarial reward to encourages the adversarial vehicle to navigate toward a potential conflict point with ego vehicle while avoiding
collisions with other surrounding vehicles, which forms the reward function as
\begin{equation}
\begin{aligned}
\mathcal{R}\left(s, a\right)=d_{\text {pos}_t}^{\text {adv }}(s_{t}, a_{t}) - d_{\text {pos}_{t-1}}^{\text {adv }}(s_{t-1}, a_{t-1})-15 r_t^{col} - 15 r_t^{fin}
\label{eq6}
\end{aligned}
\end{equation}
where $\phi_{\text {pos}_t}^{\text {adv }}(s_{t}, a_{t})$ denotes the distance between the adversarial vehicle and its goal position at time $t$, the collision cost $r_t^{col}$ is set to -1 if  adversarial vehicle collides with any other surrounding vehicles at time $t$, and the goal-reaching cost $r_t^{fin}$ is set to 1 if the adversarial vehicle is within 2 meters radius of the goal at time $t$.

\textit{2) Transition Function-based Adversarial Generation}\\
For the optimization objective, besides updating the driving policy, the transition function can also be updated.
The transition function receives the action and state to output the next traffic state, i.e., $s_{t+1} \sim f(\cdot|s_t, a_t; \boldsymbol{\theta})$.
‌Depending on the driving environment, the traffic state can encompass various elements, such as Bird's Eye View (BEV) images, LiDAR point cloud, trajectory prediction of traffic participants, high-level navigation commands, and more.
In the following, we present the transition function-based adversarial generation for two traffic state definitions, i.e., kinematics-based state and trajectory-based state.

\textbf{Kinematics-based State:}
For the simulation environment, we can denote the traffic state as $\mathbf{s}_t=\left\{p_t^i, \psi_t^i, v_t^i\right\}_{i=0}^N$, where $p_t^i$ denote the position, $\psi_t^i$ the orientation, $v_t^i$ the speed of the $i$-th agent at time $t$, $N$ the number of agents.
To unroll the simulation forward in time, we compute the state at the next time step $\mathbf{s}_{t+1}$ given the current state $\mathbf{s}_t$ and actions $\mathbf{a}_t$ of all agents using
the kinematics model $\kappa$, i.e., $\mathbf{s}_{t+1}=\kappa\left(\mathbf{s}_t, \mathbf{a}_t\right)$.
Here the transition function could be the kinematics model~\cite{hanselmann2022king}.
The kinematics model can usually be chosen as a bicycle model, which provides a strong prior on physically plausible motion of non-holonomic vehicles and is differentiable, enabling backpropagation through the unrolled state sequence. 

\textbf{Trajectory-based State:}
The other common definition of the traffic state additionally contains the future trajectory of the traffic participants~\cite{rempe2022generating,zhang2022adversarial}.
To formalize the traffic state, we denote the driving agents and represent a traffic scenario as a tuple $\mathbf{s}_t = (M,\boldsymbol{Y}_{1:t}^{ego},\left\{\boldsymbol{Y}_{1:t}^i\right\}_{i=1}^N)$ with duration $t$ time steps, in which the high-definition road map M consists of road shapes, traffic signs, traffic lights, etc, $\boldsymbol{Y}_{1:t}^{ego}$ and $\left\{\boldsymbol{Y}_{1:t}^i\right\}_{i=1}^N = \left[Y_{1:t}^{1}, \ldots, Y_{1:t}^{N}\right]$ denote the history trajectory of ego vehicle and sounding vehicles, respectively.
The fundamental problem is to update trajectory prediction model by generating compliant future traffic trajectories of adversarial agent that are prone to collisions with the ego agent’s rollouts. 
We define a binary random variable $Coll=\{Coll_T, Coll_F\}$
to denote whether ego agent collides with adversarial agents.
Formally, $\boldsymbol{Y}_{t:T}^{ego}$ and $\boldsymbol{Y}_{t:T}^{adv}$ stand for the future trajectories of ego vehicle and sounding vehicles starting from step $t$, respectively.
Thus, through modeling trajectory probability distribution between ego agent and adversarial agents, the adversarial scenario generation is expressed as
\begin{equation}
\min _{f} J\left(\pi, f; \boldsymbol{\theta}\right) \Leftrightarrow \max _{\boldsymbol{Y}^{adv}} \sum_{} \mathbb{P}\left(Y^{ego}, \boldsymbol{Y}^{adv} \mid Coll_T, \mathbf{s}_t\right).
\label{eq7}
\end{equation}
under the assumptions that the ego vehicle’s reactions are unidirectionally based on the future traffic, we can factorize the right term of Eq.\ref{eq7} with the Bayesian formula 
\begin{small} 
\begin{equation}
\resizebox{.98\linewidth}{!}{$
    \displaystyle
\begin{aligned}
& \max _{\boldsymbol{Y}^{adv}} \sum_{} \mathbb{P}\left(Y^{ego}, \boldsymbol{Y}^{adv} \mid Coll_T, \mathbf{s}_t\right) \\
& \propto \max _{\boldsymbol{Y}^{adv}} \underbrace{\mathbb{P}\left(\boldsymbol{Y}^{adv} \mid \mathbf{s}_t\right)}_{\text { 1st Term }} \sum_{} \underbrace{\mathbb{P}\left(Y^{ego} \mid \boldsymbol{Y}^{adv}, \mathbf{s}_t\right)}_{\text {2nd Term }} \underbrace{\mathbb{P}\left(Coll_T \mid Y^{ego}, \boldsymbol{Y}^{adv}\right)}_{\text {3rd Term }}. \\
&
\end{aligned}$}
\label{eq8}
\end{equation}
\end{small} 
It is beneficial to perform the above scenario generation probability factorization since each term in
Eq.\ref{eq8} features a specific meaning and is tractable to handle.
The 1st term is the standard trajectory prediction problem. The 2nd term denotes the interactive ego trajectory yielding to the current state
and upcoming traffic flow, which is conditioned on the driving policy $\pi$. 
The 3rd term reflects the likelihood of a collision in the compositional future, which can be simulated directly or treated as a binary classifier to fit. 
Thus, it is possible to approach the near-optimal adversarial trajectory via updating trajectory prediction model while fixing driving policy models and binary classifier.

\section{AI Component-level Evaluation on Safety-critical Scenarios}
Object detection is an important task in autonomous driving to perceive the surroundings. 
We evaluate the safety and robustness of object detection on natural driving scenarios and safety-critical scenarios, where safety-critical scenarios include manual adversarial attacks and natural distribution shift.
First, we evaluated the performance of the object detection models against digital attacks on commonly used open-loop driving dataset.
Then, we assess the safety of object detection models to physical attack algorithms in the CARLA physical simulation environment.
Furthermore, we developed a toolbox capable of transforming natural driving data into safety-critical scenario with distribution shifts, and evaluated the performance of object detection under these generated safety-critical scenarios.


\subsection{AI Component-level Evaluation for Adversarial Attacks}
For adversarial attacks, we first evaluate the performance of digital attacks on open-loop autonomous driving dataset.
Then, we evaluate the performance of physical attacks on 3D object detection in the simulation of physical world.

\textit{1) Adversarial Attacks in Digital World}

\textbf{Dataset:} For digital attacks, we use commonly-used AD dataset nuScenes~\cite{caesar2020nuscenes}.
The nuScenes dataset contains 1000 sequences of approximately 20s duration with a LiDAR frequency of 20
FPS. The box annotations are provided for every 0.5s.
Each
frame has one point cloud and six images covering $360^\circ$
horizontal FOV.
In total, there are 40k frames which are
split into 28k, 6k, 6k for training, validation, and testing. 

\textbf{Object Detection Models and Attack Baselines:}
To evaluate the performance of 3D object detection on digital adversarial attacks, we select 2 representative 3D object detection models with open source code, FCOS3D and PGD\footnote{In the following, we use the notation PGD-Det to distinguish between the object detection method PGD and the adversarial attack method PGD.} models.
We implement these object detection models on the open source framework MMDetection3D~\cite{mmdet3d2020}.

Regarding the adversarial attack baselines, we adopt 3 typical gradient-based attack methods, Fast Gradient Sign Method (FGSM)~\cite{goodfellow2014explaining}, Projected Gradient Descent (PGD)~\cite{madry2018towards} and Carlini-Wagner (C$\&$W)~\cite{carlini2017towards}, to generate $L_{\infty}$ adversarial perturbations for each object detection model.
Our experiment setup fixes the maximum perturbation value at $\epsilon = 5$. The number of iterations for PGD is set to 10.
The
process begins with the introduction of Gaussian noise to randomly perturb input images.
We further investigate the effect of varying the budget of perturbation $\epsilon$ from 0 to 8 on the performance in the experimental analysis.


\textbf{Evaluation Metrics:}
For 3D object detection, the main evaluation metrics are mean Average Precision (mAP) and nuScenes detection score (NDS) computed on 10 object categories. The mAP
is calculated using the 2D center distance on the ground
plane instead of the 3D IoU. The NDS metric consolidates
mAP and other aspects (e.g., scale, orientation) into a unified score. 

\begin{table}[t]
\small
\caption{The mAP and NDS of different 3D object detection models under digital attacks on nuScenes dataset.}
\centering
{\resizebox{\linewidth}{!}{
\begin{tabular}{cccccc}
\toprule
\multicolumn{2}{c}{\multirow{2}{*}{Adversarial attack}}&\multicolumn{2}{c}{FCOS3D}&\multicolumn{2}{c}{PGD}\\
&&mAP&NDS&mAP&NDS\\
\midrule
\multicolumn{2}{c}{Benign}&29.8&37.7&31.74&39.34\\
\midrule
\multirow{3}{*}{Digital attacks}&PGD&3.90&18.73&19.61&31.42\\
&FGSM&8.30&21.44&8.77&17.99\\
&C$\&$W&2.59&7.18&2.71&8.28\\
\bottomrule
\end{tabular}
}}
\label{Digital_1}
\end{table}

\begin{figure}[t] 
\centering
\captionsetup{font={footnotesize}}
\subfigure{
\centering
\includegraphics[width=0.48\linewidth,scale=1.00]{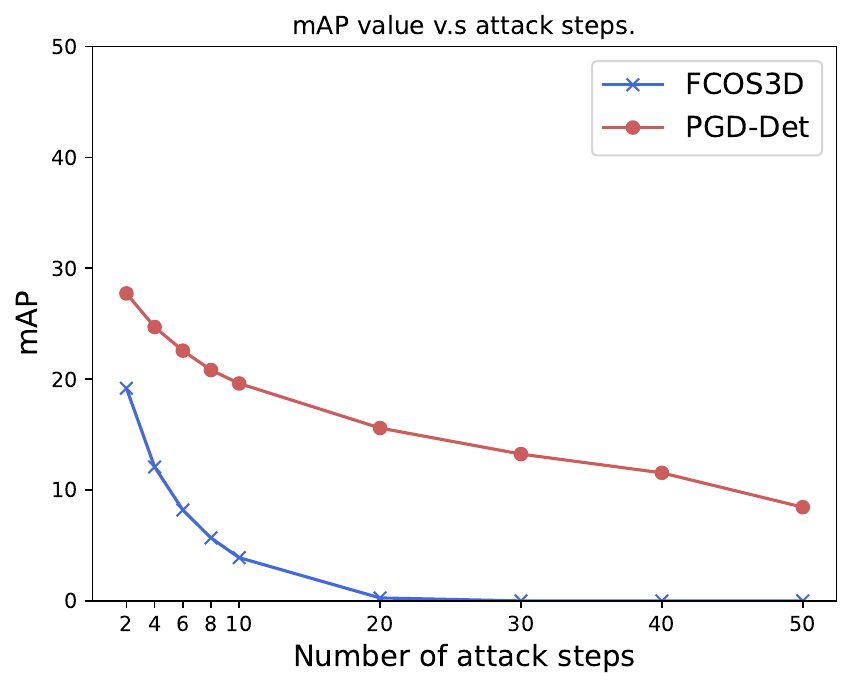}
}\hspace{-3mm}
\subfigure{
\centering
\includegraphics[width=0.48\linewidth,scale=1.00]{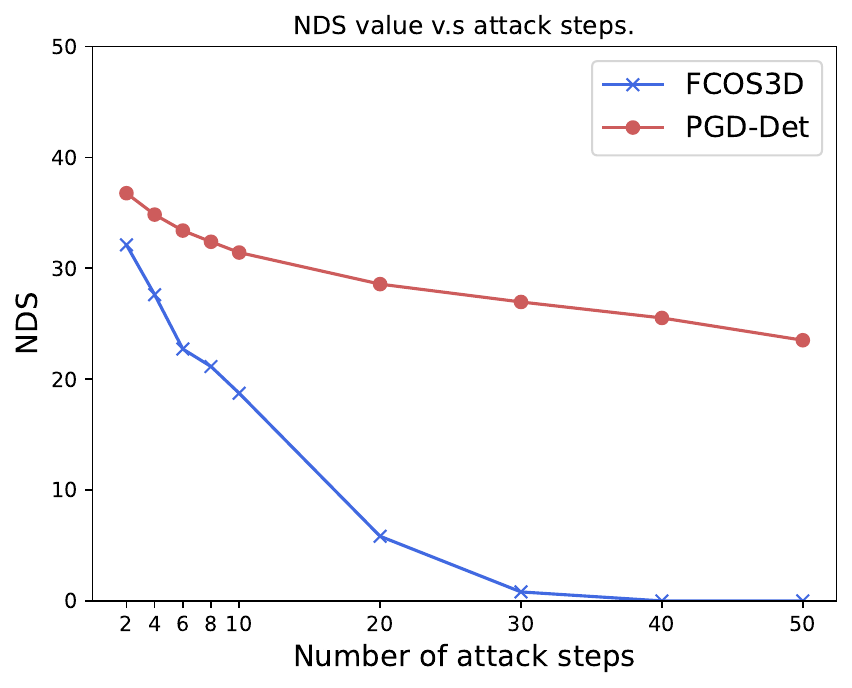}
}
\caption{mAP and NDS v.s attack iterations.}
\label{figure:4}
\end{figure}

\textbf{Experimental Results:}
Tab.\ref{Digital_1} presents the experimental results of various digital attacks on two 3D object detection models. 
The mAP of the FCOS3D model on the benign data is 29.8, which drops to 3.9, 8.3 and 2.6 under attacks from PGD, FGSM and C$\&$W, respectively.
A similar phenomenon is observed with the object detection model PGD-Det.
Additionally, we observe that the attack effect of PGD on the PGD-Det model is less effective compared to FCOS3D model.
Therefore, in experimental analysis, we further explore the impact of different parameter settings of the PGD attack method on the performance of the PGD-Det model.

\textbf{Experimental Analysis:}
We show the curves of mAP and NDS of different object detection models under PGD digital attack by varying the iteration steps.
Overall, as the iteration step increases, the model is more affected by the PGD and the the worse the performance becomes.
There is a difference in the impact of the different object detection models on PGD method, and the performance of the FCOS3D decreases more significantly than that of PGD-Det.

\textit{2) Adversarial Attacks in Simulation of Physical World}

\textbf{Datasets:} Since the objects of open-loop datasets are not 3D assets, i.e., the texture of the object is not differentiable, the camouflage-based physical attacks cannot be deployed on these datasets. 
Thus, we select the simulator CARLA~\cite{dosovitskiy2017carla}, a prevalent open source simulator for autonomous driving research, as our
3D simulator. 
For the training of physical attacks, we collect training data in the CARLA map.
Unlike the previous work~\cite{zhang2023boosting} which randomly choose spawn locations for target vehicle and does not follow the coherence of the vehicle driving, we set the start and end points and execute the Autopilot or control algorithm to drive vehicle. 
Then the camera is spawned at a distance of 20 meters behind the target vehicle and acquires a frame at an interval of time.
To enrich the diversity of the driving data, we collect two training datasets in two CARLA maps, containing 400 and 600 images, respectively, named Town05 and Town06.
For the test of physical attacks, unlike previous work~\cite{zhang2023boosting} that separates a portion of the collected data into a test set for offline test, our evaluation supports real-time test in CARLA simulation.
The adversarial patch/texture trained on Town06 are tested on Town04, trained on Town05 are tested on Town03. Each frame is tested in CARLA map during the driving of target vehicle and a total of 400 frames are evaluated.

\textbf{Object Detection Models and Attack Baselines: }
We use YoloV3 and YoloV5 as the object detection models.
Due to some gaps between the simulated CARLA data and the real-world data, it leads to the open-sourced YoloV3 and YoloV5 models being inferior performance on the CARLA data.
Thus, prior to conducting the attack, we train YoloV3 and YoloV5 using the CARLA data, and then the pre-trained YoloV3 and YoloV5 are used to evaluate the performance of object detection under physical attacks.
The model is initialized with the pre-trained weights from the well-known COCO benchmark~\cite{lin2014microsoft}.
Then the model is finetuned on the CARLA data using SGD optimizer with a momentum of 0.99 and a weight decay of 4e-4. The learning rate is linearly decreased from 0.01 to 0.001.

For the patch-based physical attacks, we adopt the Adv. Patch method~\cite{shrestha2023towards} to generate a adversarial patch on the target victim vehicle.
We follow the settings of original paper to optimize the adversarial loss function using the Adam optimizer with a learning rate of 0.04, an $\epsilon$ of 1e-8 for 200 epochs.
For camouflage-based physical attacks, we adopt DAS~\cite{wang2021dual} and TPA~\cite{zhang2023boosting} methods to generate the adversarial camouflage, which adopt a SGD optimizer with a learning rate of 0.01, a weight decay of 1e-4, and a maximum of 200 epochs.

\textbf{Evaluation Metrics: }
We use AP (0.5) as the evaluation metric, which is Average Precision (AP) considering an Intersection over Union (IoU) threshold of 0.5, to evaluate the performance of class ``Car''.
Apart form AP (0.5), we also select another metric, Attack Success Rate (ASR), by calculating the ratio $(n_{0}-n_{1})/n_{0}$, where $n_{0}$ or $n_{1}$ denotes the number of detected vehicle in benign or adversarial images. 

\begin{table}[t]
\small
\caption{The mAP(0.5) and NDS of object detection models on Town04 dataset against different physical attacks trained with Town06 dataset.}
\centering
{\resizebox{\linewidth}{!}{
\begin{tabular}{cccccc}
\toprule
\multicolumn{2}{c}{\multirow{2}{*}{Adversarial attack}}&\multicolumn{2}{c}{YoloV5}&\multicolumn{2}{c}{YoloV3}\\
&&AP (0.5)&ASR&AP (0.5)&ASR\\
\midrule
\multicolumn{2}{c}{Benign}&99.8&-&99.5&-\\
\midrule
\multirow{2}{*}{Patch-based attack}&Random&24.3&75.4&45.4&66.8\\
&Adv. Patch&16.5&86.5&36.0&83.3\\
\midrule
\multirow{2}{*}{Camouflage-based attack}&Random&98.2&-&99.3&-\\
&DAS&56.2&80.3&65.8&74.2\\
&TPA&62.5&75.1&63.9&76.3\\
\bottomrule
\end{tabular}
}}
\label{physical1}
\end{table}

\begin{table}[t]
\small
\caption{The mAP(0.5) and NDS of object detection models on Town03 dataset against different physical attacks trained with Town05 dataset.}
\centering
{\resizebox{\linewidth}{!}{
\begin{tabular}{cccccc}
\toprule
\multicolumn{2}{c}{\multirow{2}{*}{Adversarial attack}}&\multicolumn{2}{c}{YoloV5}&\multicolumn{2}{c}{YoloV3}\\
&&AP (0.5)&ASR&AP (0.5)&ASR\\
\midrule
\multicolumn{2}{c}{Benign}&99.1&-&99.3&-\\
\midrule
\multirow{2}{*}{Patch-based attack}&Random&58.2&64.2&95.9&65.8\\
&Adv. Patch&37.2&74.7&73.6&76.1\\
\midrule
\multirow{2}{*}{Camouflage-based attack}&Random&95.6&8.7&99.5&-\\
&DAS&49.5&91.5&54.9&88.3\\
&TPA&50.6&87.8&56.9&86.7\\
\bottomrule
\end{tabular}
}}
\label{physical2}
\end{table}

\textbf{Experimental Results: }
Tabs.\ref{physical1} and \ref{physical2} quantitatively report the results of the two object detection models under different adversarial attack methods, respectively. In benign case, the performance of the pre-trained object detection models YoloV3 and YoloV5 can achieve an AP(0.5) of more than 99\% on Town03 and Town04.
For the patch-based attack, random patches can lead to wrong detection of the object detection models to some extent. Furthermore, adversarial patches can produce a more effective attack on the object detection models.
For the camouflage-based attack, the vehicle surface texture initialized with the random noise has little effect on the performance of the object detection models.
On the contrary, the DAS and TPA methods can cause the object detection model to produce wrong detection by generating the adversarial texture.
Overall, it can be seen that the camouflage-based attack is not as effective as the patch-based attack, but the camouflage-based methods can attack the object detection model more stealthily.
\begin{figure*}[ht]
\centering     
\subfigure[Benign]{\label{fig3:0}\includegraphics[width=0.32\linewidth]{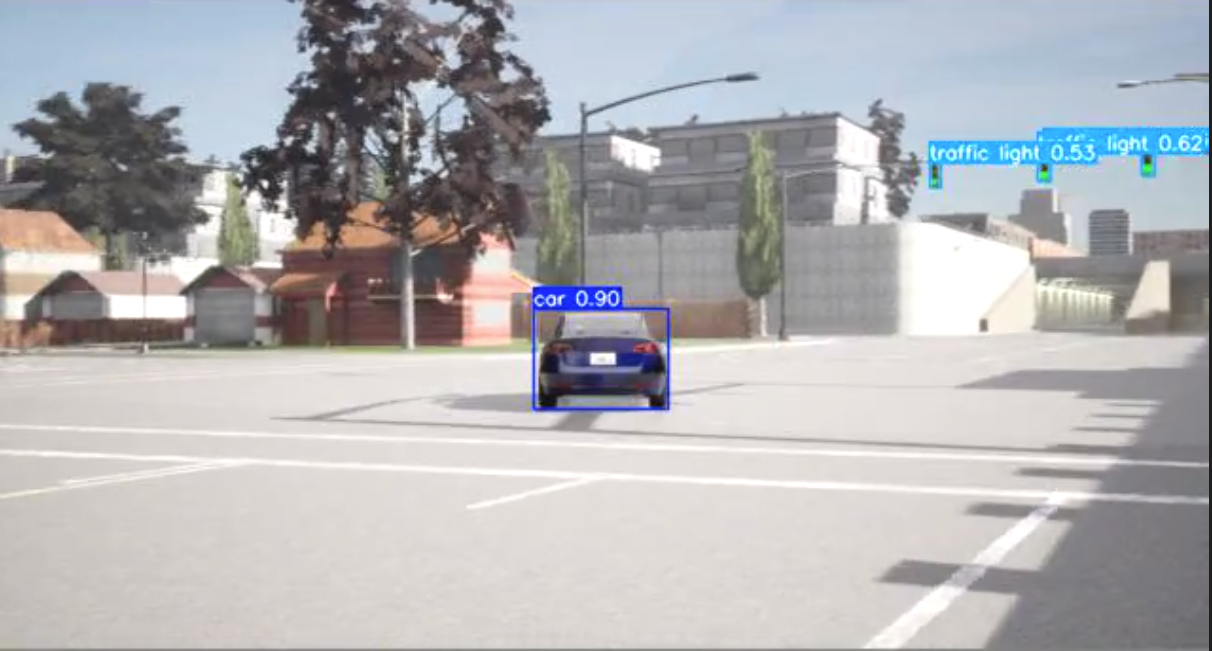}}
\subfigure[Random Patch]{\label{fig3:0}\includegraphics[width=0.32\linewidth]{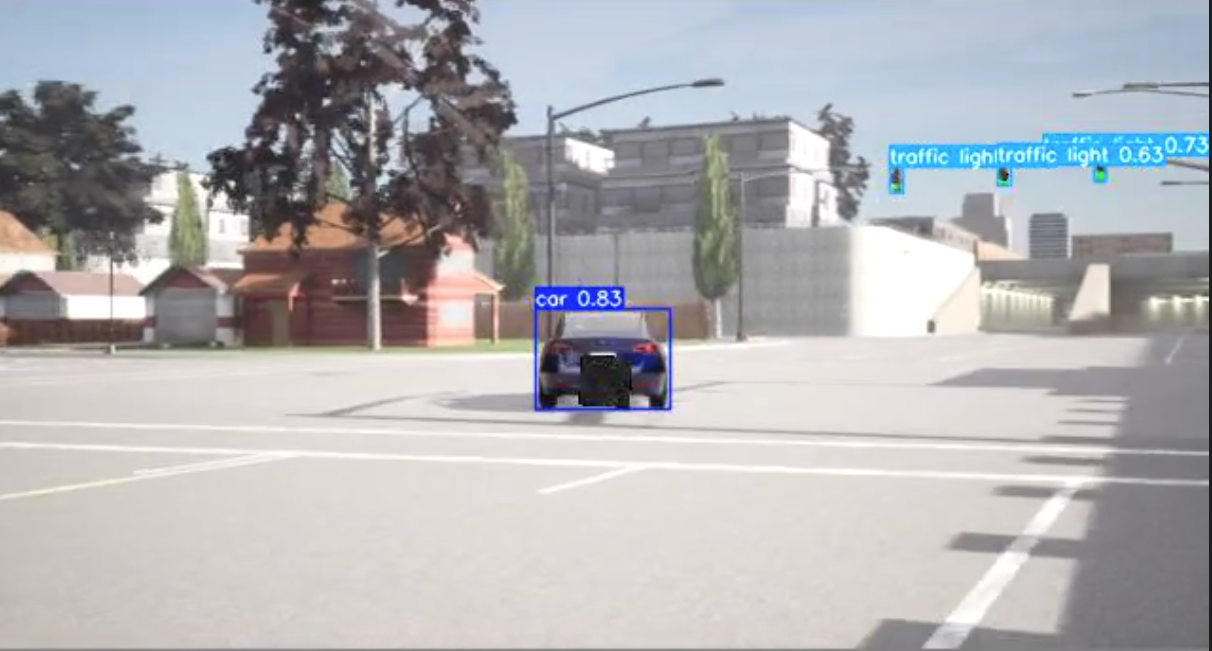}}
\subfigure[Adv. Patch]{\label{fig3:1}\includegraphics[width=0.32\linewidth]{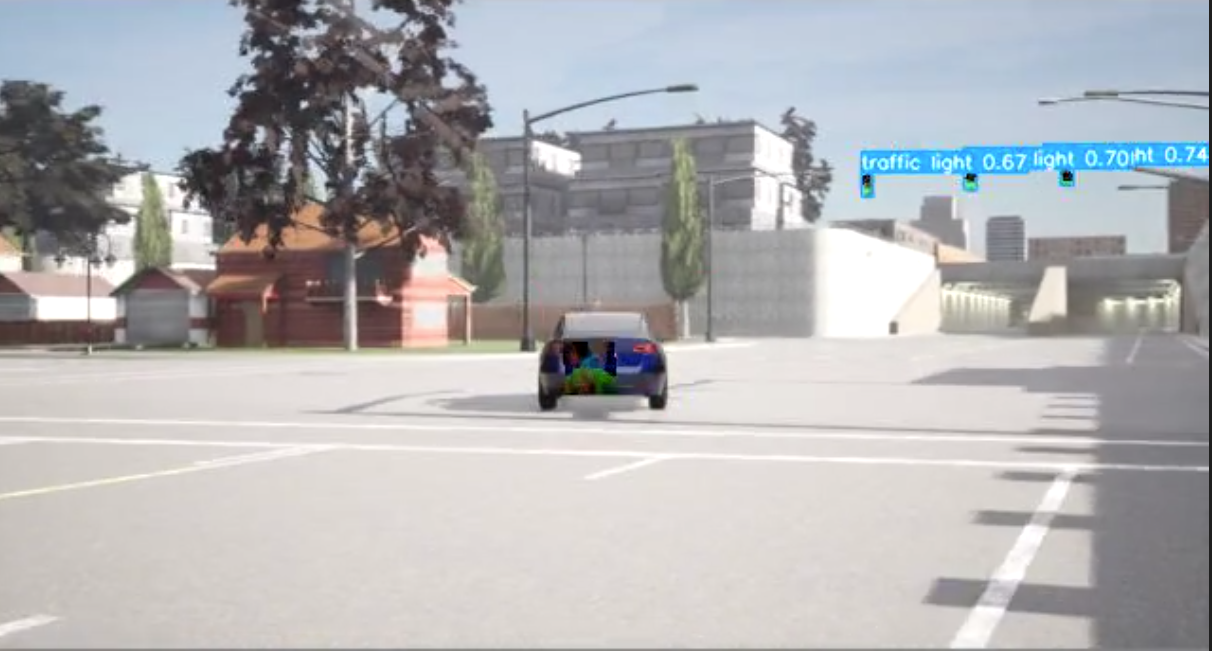}}
\subfigure[Random Camouflage]{\label{fig3:2}\includegraphics[width=0.315\linewidth]{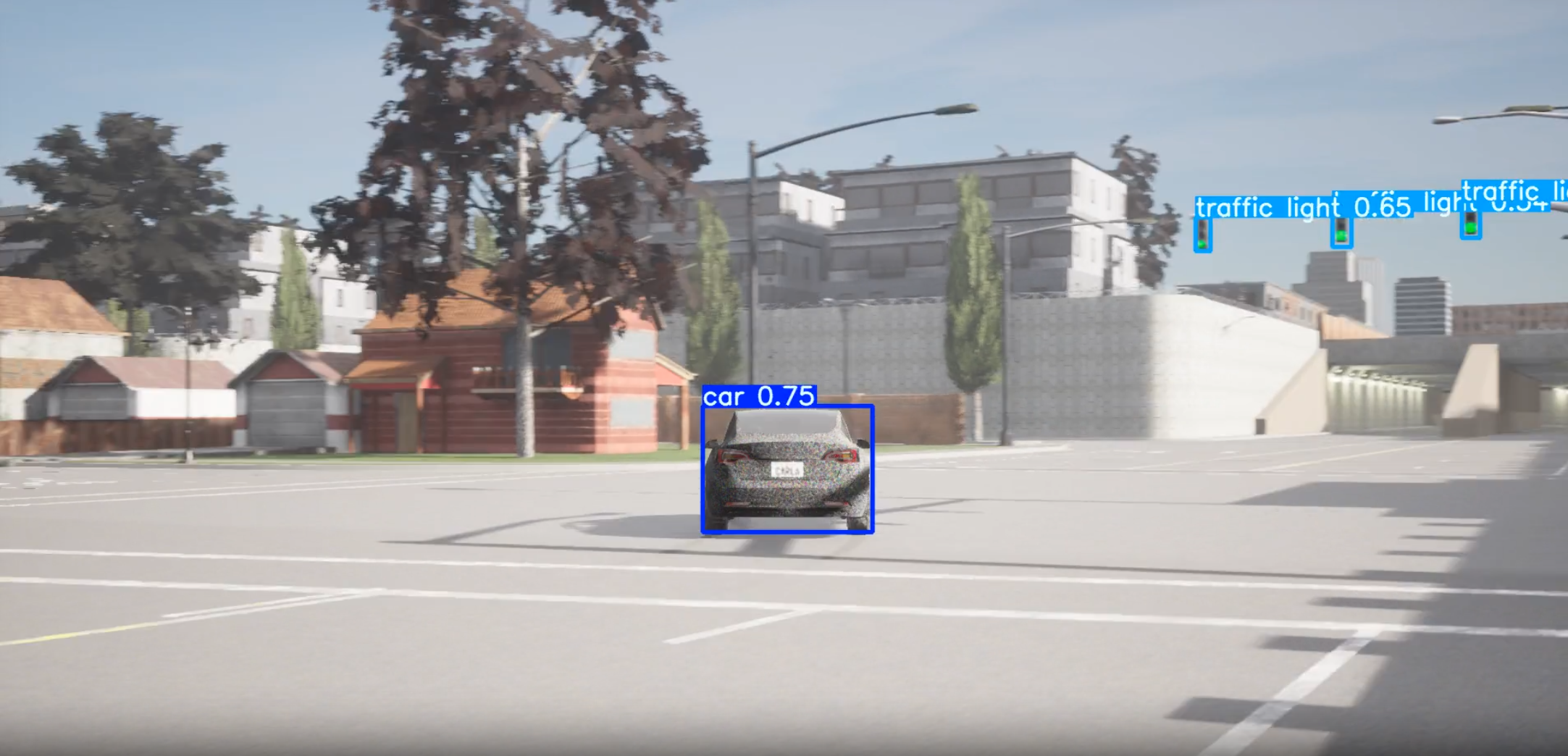}}
\subfigure[DAS]{\label{fig3:3}\includegraphics[width=0.32\linewidth]{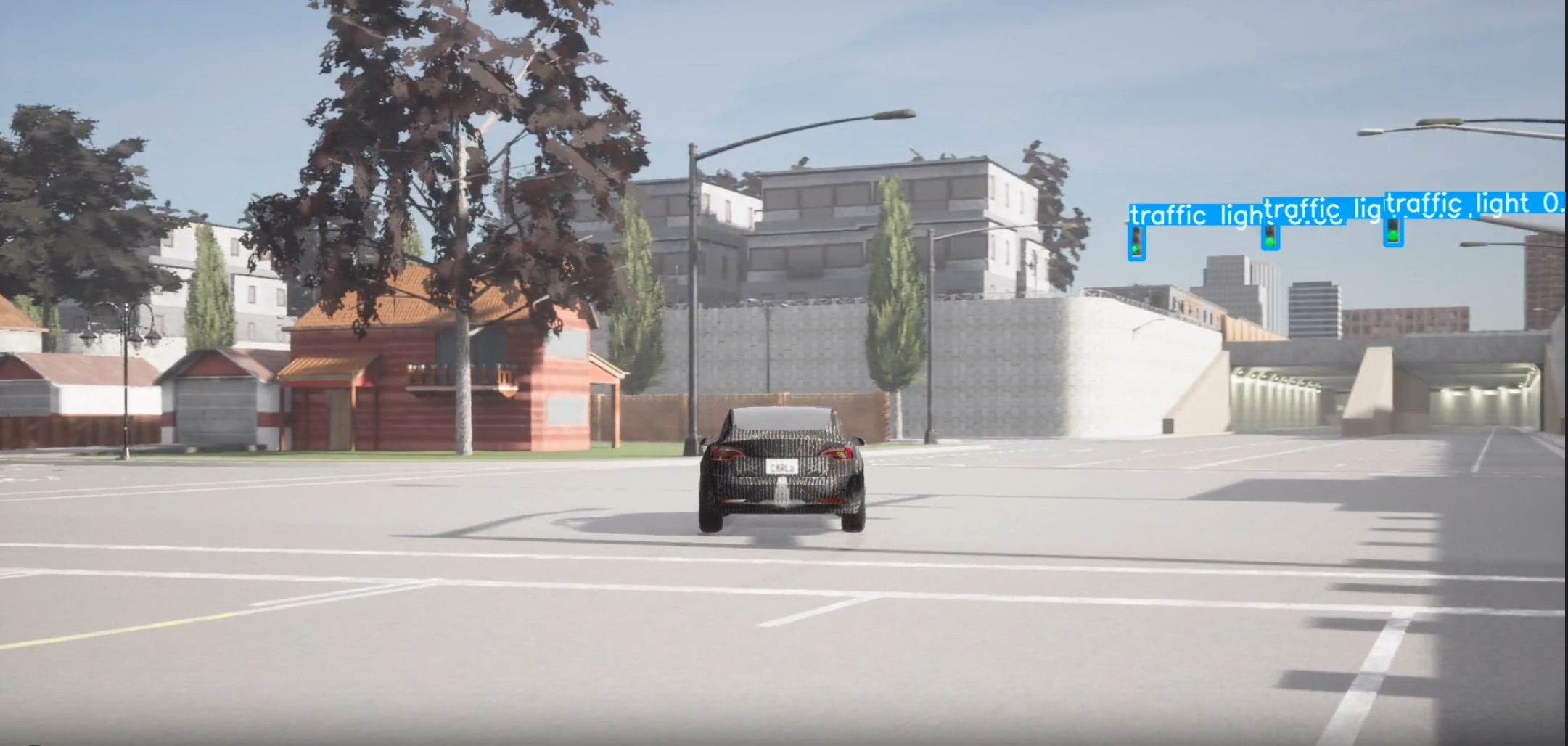}}
\subfigure[TPA]{\label{fig3:4}\includegraphics[width=0.32\linewidth]{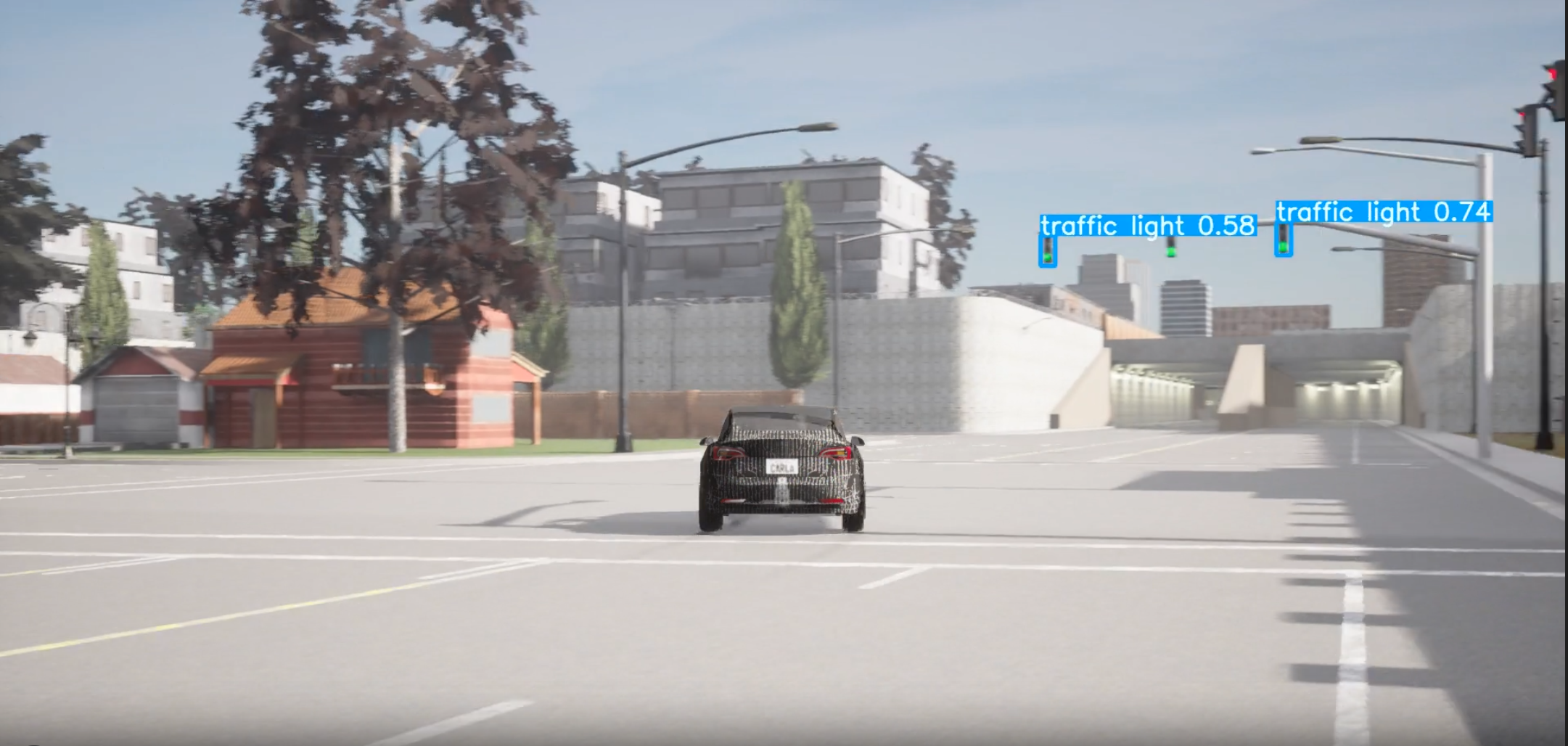}}
\caption{Visualization of object detection results of various physical attack algorithms on Town03 versus randomly generated patches and textures on YoloV5 model.}
\label{figure_vis}
\end{figure*}

\textbf{Experimental Analysis: }
As shown in Fig.\ref{figure_vis}, we investigate the effect of various physical attack methods on object detection model YoloV5 deployed in the simulation environment and use Tesla Model 3 as the  target victim vehicle. Then we show the detection results of the object detection models on the victim vehicle.
For the patch-based attack, the object detection can effectively attack the object detection model leading to wrong detection results, e.g., the YoloV5 model does not recognize the target vehicle in the adversarial attack scenario yet can accurately recognize the traffic sign.
For camouflage-based attack, the object detection model can recognize a vehicle covered with a random noise texture. However, DAS and TPA with adversarial texture can make the object detection model fail to recognize the victim vehicle while the camouflage-based attack is more stealthy.

%
\subsection{AI Component-level Evaluation for Natural Distribution Shift}
\begin{figure*}[ht]
\centering     
\subfigure[Origin]{\label{fig3:0}\includegraphics[width=0.23\linewidth]{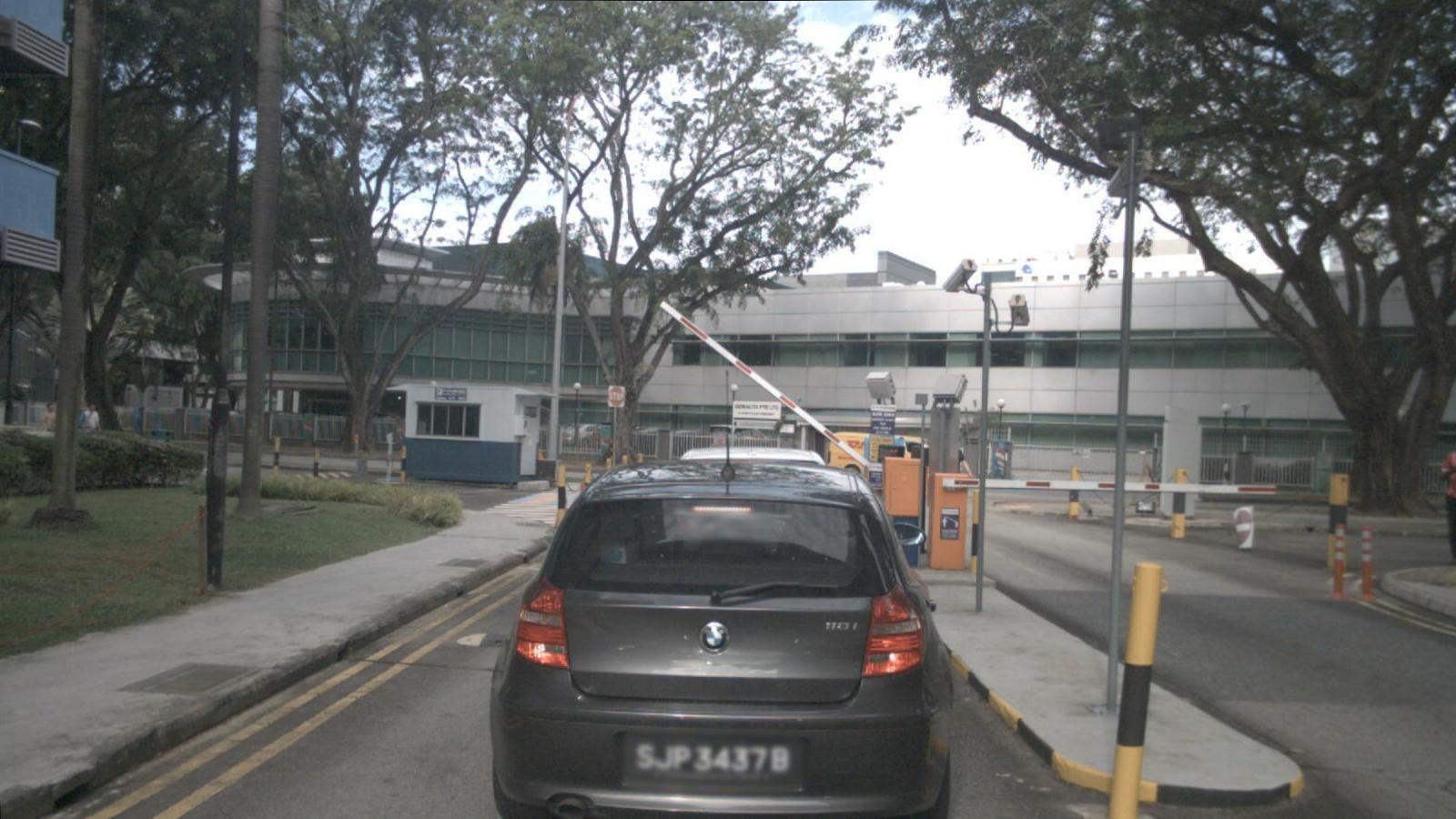}}
\subfigure[Snow]{\label{fig3:0}\includegraphics[width=0.23\linewidth]{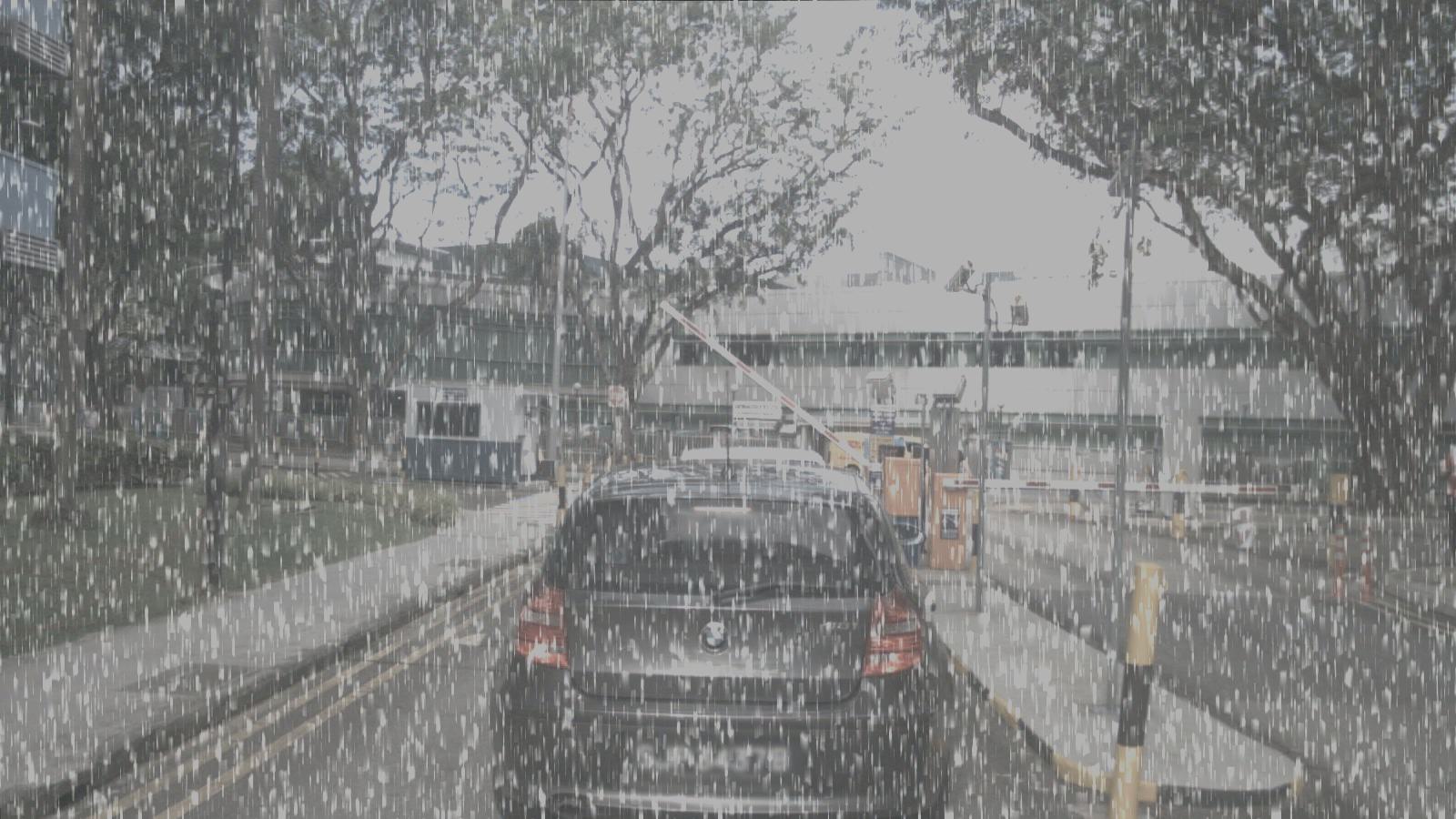}}
\subfigure[Rain]{\label{fig3:1}\includegraphics[width=0.23\linewidth]{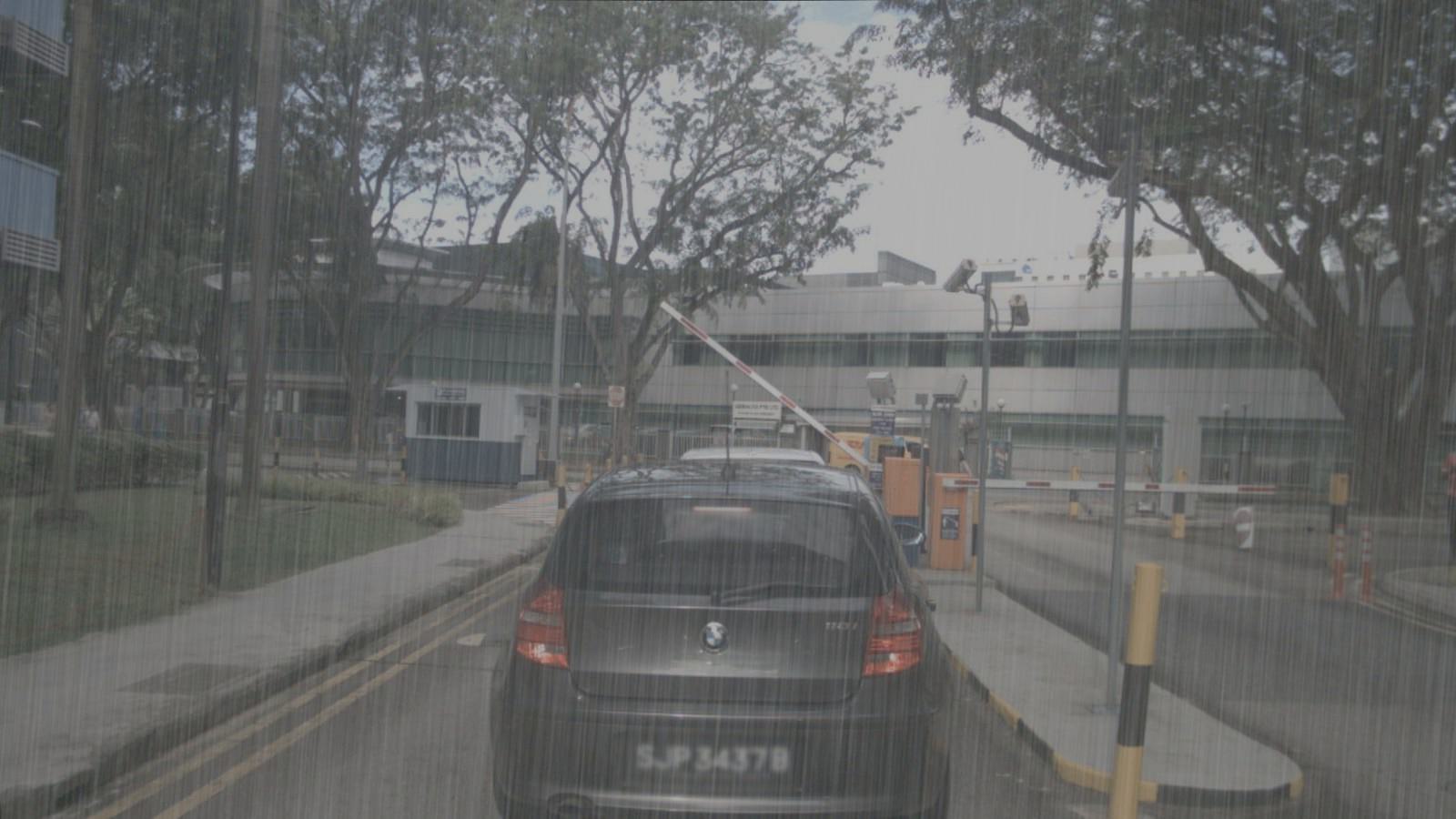}}
\subfigure[Fog]{\label{fig3:2}\includegraphics[width=0.23\linewidth]{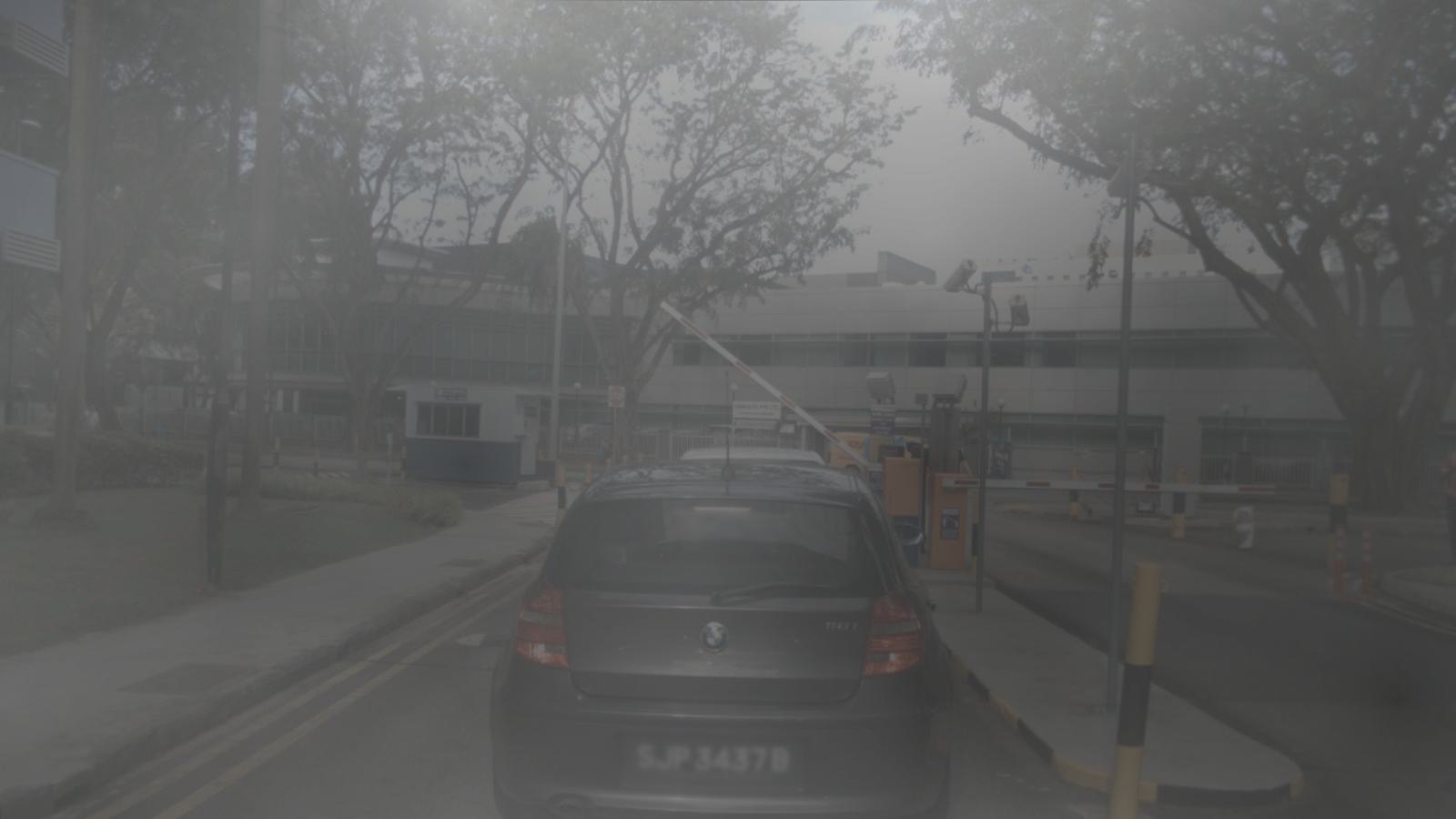}}
\subfigure[Sunlight]{\label{fig3:3}\includegraphics[width=0.23\linewidth]{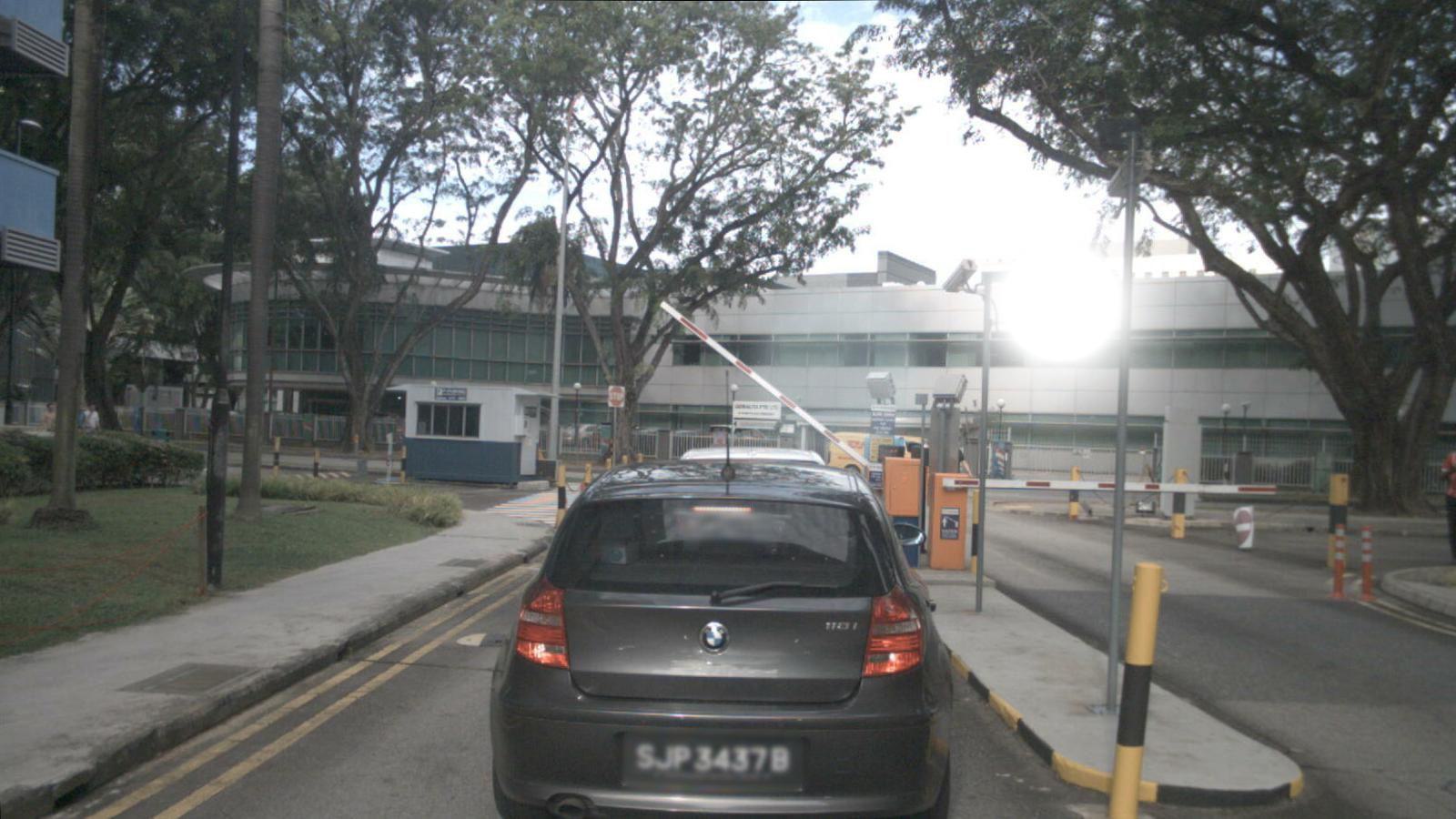}}
\subfigure[Uniform Noise]{\label{fig3:4}\includegraphics[width=0.23\linewidth]{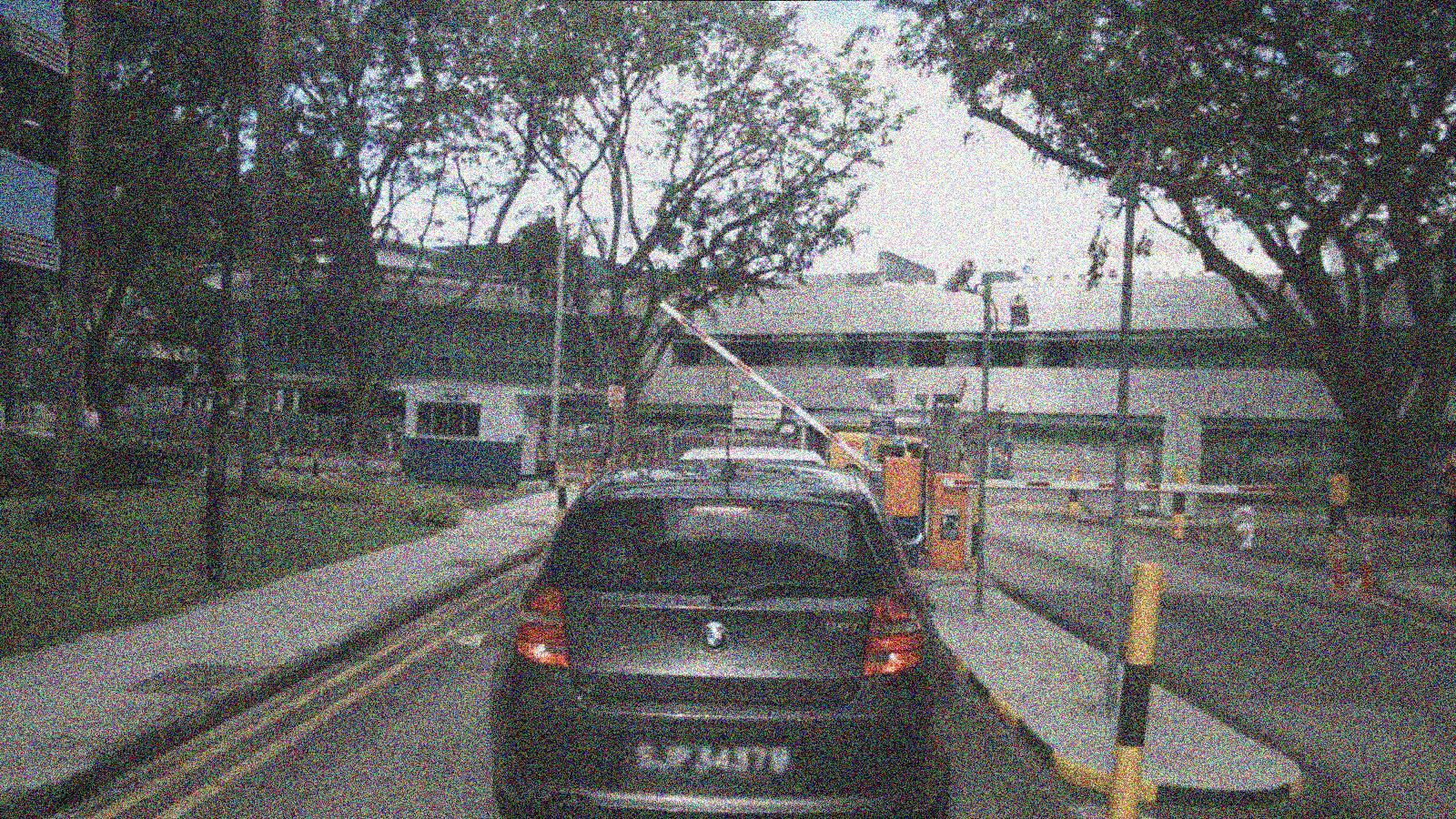}}
\subfigure[Gaussian Noise ]{\label{fig3:5}\includegraphics[width=0.23\linewidth]{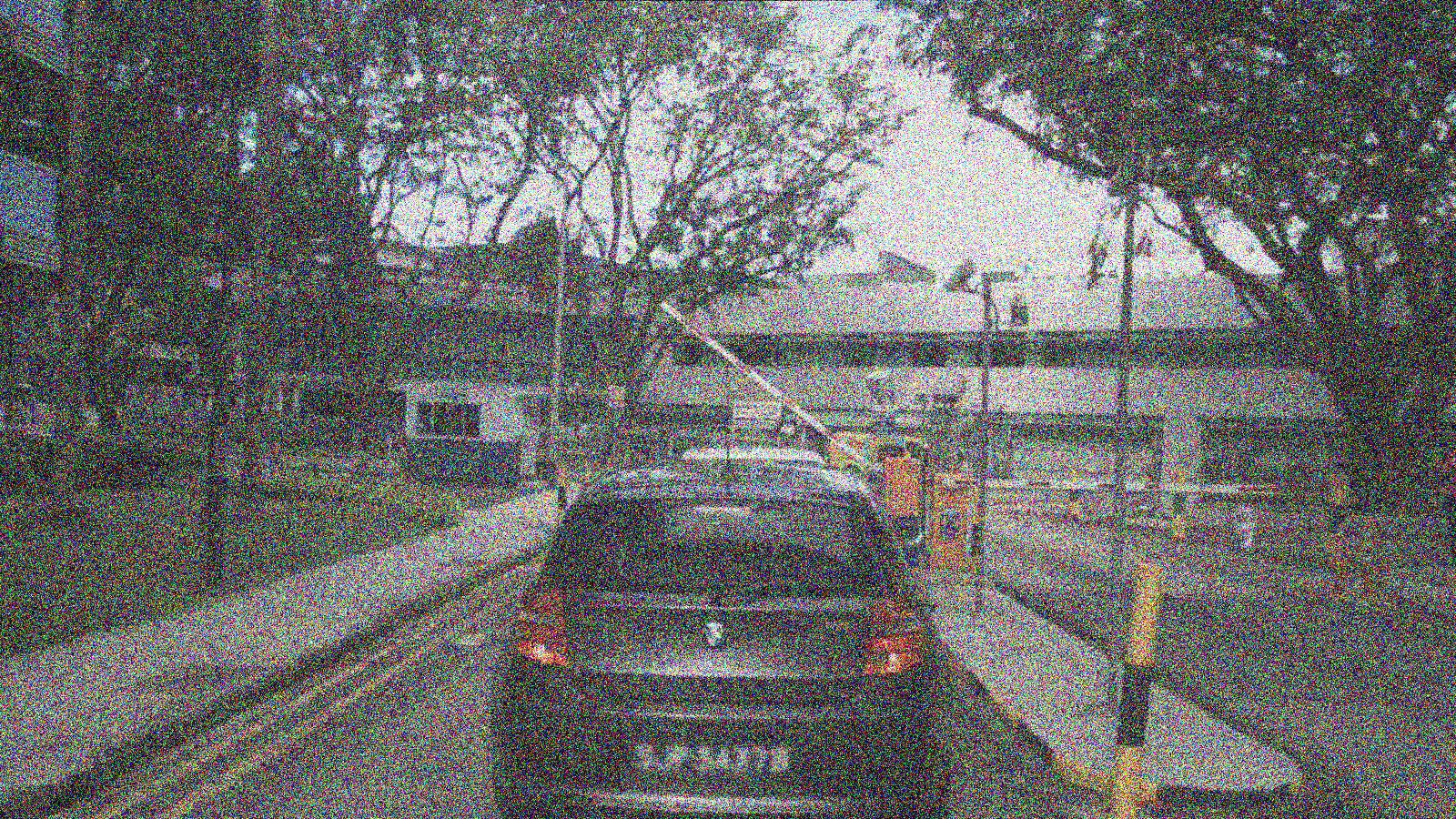}}
\subfigure[Impulse Noise]{\label{fig3:6}\includegraphics[width=0.23\linewidth]{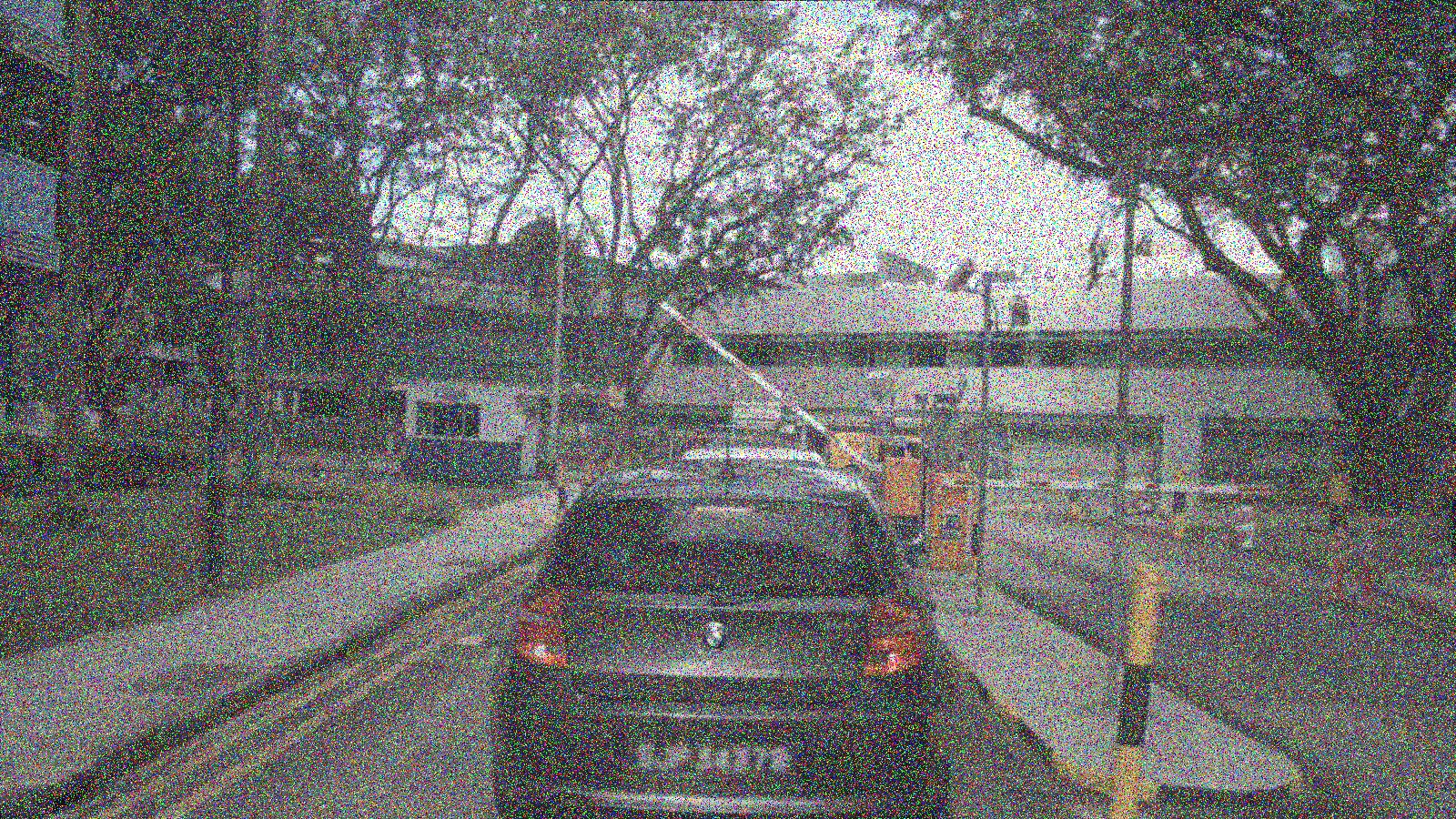}}
\subfigure[Scale]{\label{fig3:7}\includegraphics[width=0.23\linewidth]{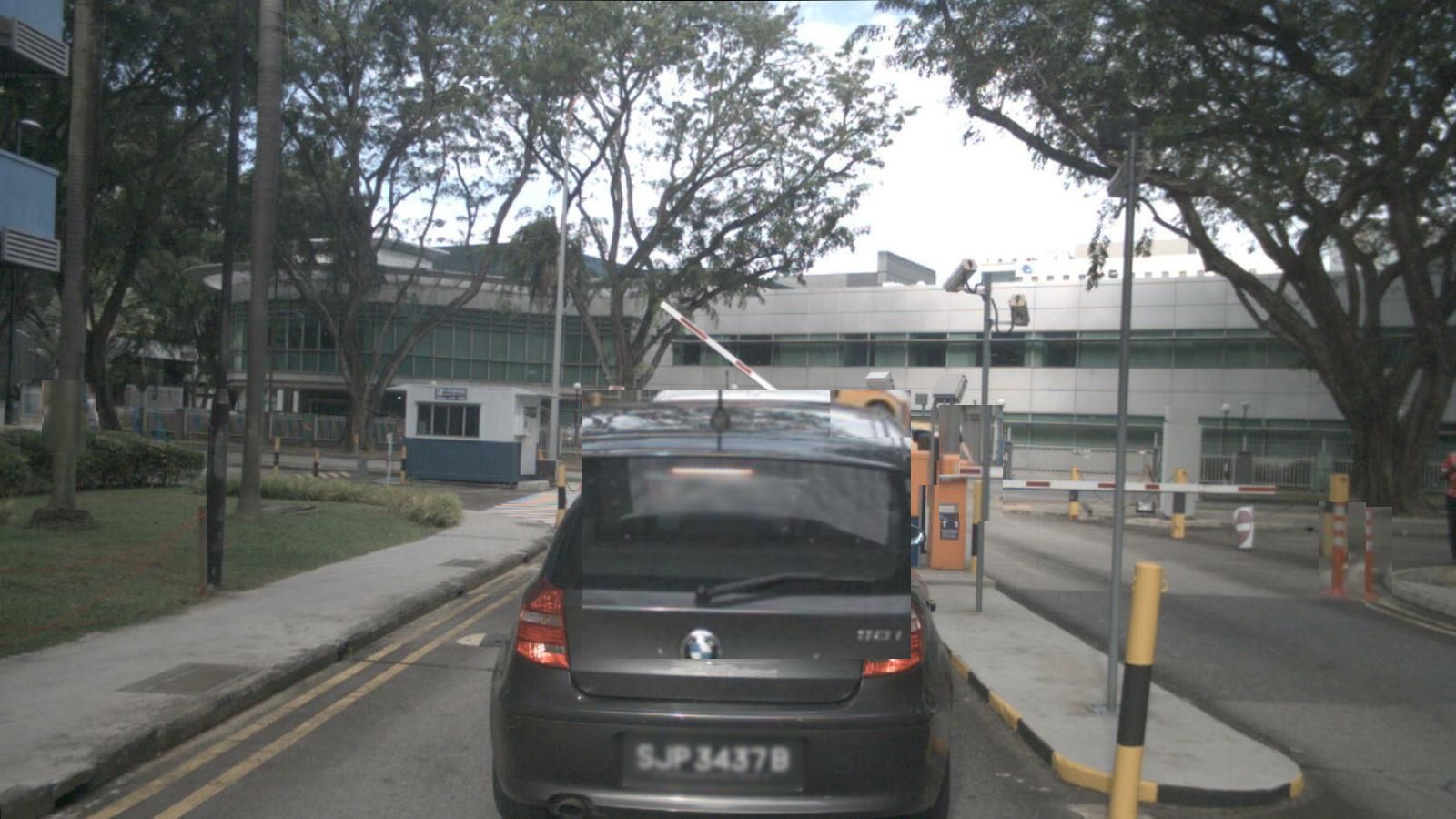}}
\subfigure[Shear]{\label{fig3:8}\includegraphics[width=0.23\linewidth]{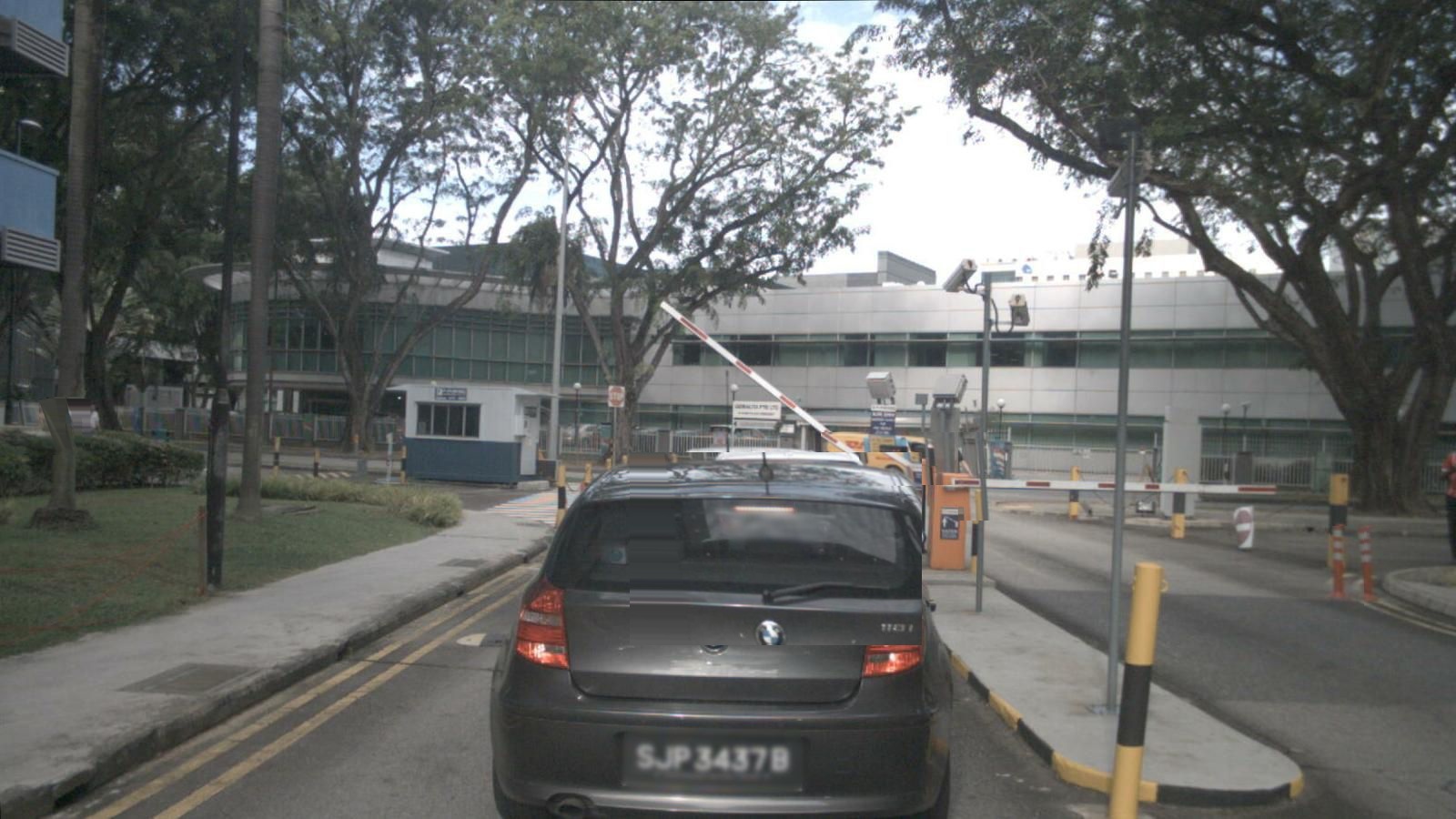}}
\subfigure[Rotation]{\label{fig3:9}\includegraphics[width=0.23\linewidth]{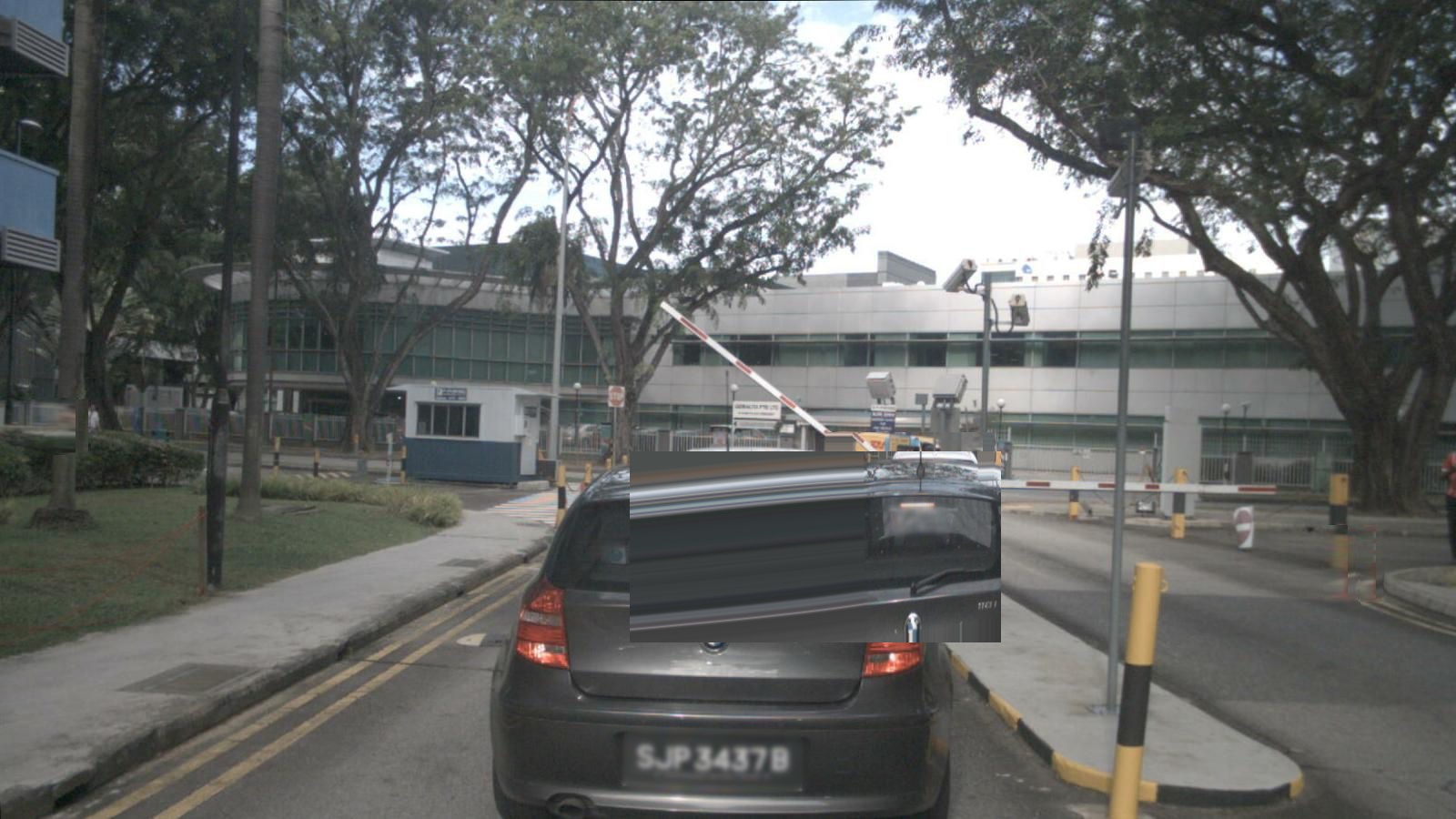}}
\subfigure[Motion Blur]{\label{fig3:10}\includegraphics[width=0.23\linewidth]{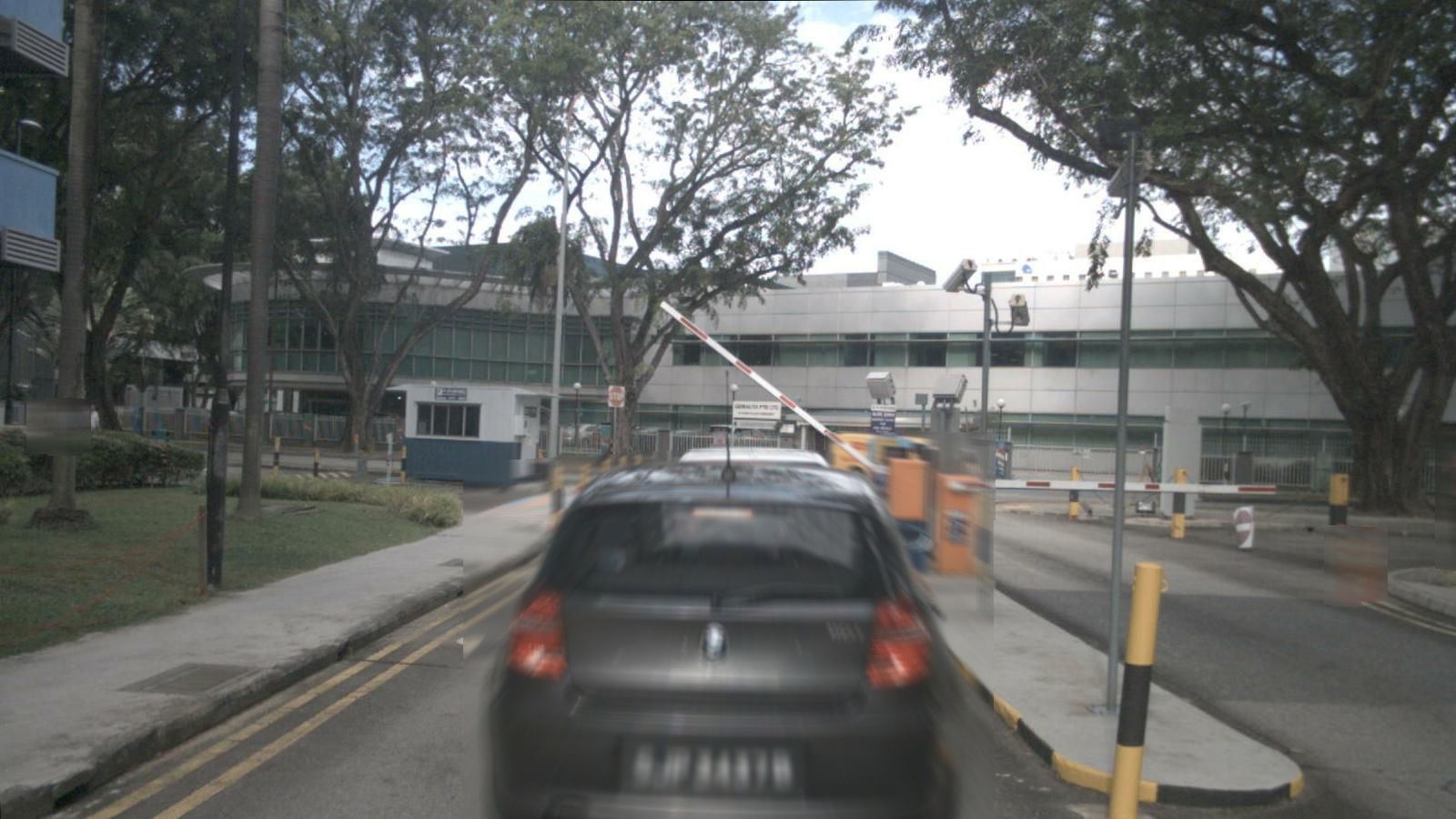}}
\caption{The original sample and its
noisy version injected with the various types of distribution shift including environment noise, sensor noise and object noise.}
\label{distributionshift}
\end{figure*}

In addition to the manually adversarial attack scenarios, we further evaluate the robustness of the AI component-level on driving scenarios with natural distributional shift.
As aforementioned in Sec.\ref{sec3}, based on driving
dataset nuScenes, we construct a benchmark nuScenes-N with distribution shift which includes environment noise, sensor noise, and object noise.
Specifically, we apply the distribution shift toolbox to inject noise into the validation set of nuScenes and obtain dataset nuScenes-N.
In Fig.\ref{distributionshift}, we show the original sample and its noisy version injected with various types of distribution shift.
We divide the injected noise strength into five levels from weak to strong allowing flexibility in controlling the strength of the distributional shift.

Subsequently, we evaluate the robustness of the 2 3D object detection models, FCOS3D and PGD-Det, on a clean dataset nuScenes and a dataset nuScenes-N containing distribution shift, respectively.
In the experiments, the strength of each type of distribution shift is set to a random value ranging from strength 1 to 5. We use mAP and NDS as evaluation metrics. The
results are evaluated based on the car class.

The experimental results are shown in Tab.\ref{table6}.
Above all, the performance of object detection model is negatively affected to different degrees when there are various distributional shift in the driving scenarios.
Specifically, for environment noise, snow can have a serious impact on the object detection models, with only 2\% and 3\% of mAP for FCOS3D and PGD-Det.
Similarly, the sensor noise can also lead to a substantial performance degradation.
Therefore, although the current open source AI models report impressive results on public benchmarks, the models will inevitably encounter distributional shift in practical applications, and these distributional shift will lead to degradation of model performance.
Therefore, designing robust AI models in practical applications is an urgent need, especially in applications such as AD, which requires a high degree of safety and robustness.

\begin{table}[t]
\small
\caption{The results of 3D object detection on nuScenes with different types of distributional shift.}
\centering
{\resizebox{\linewidth}{!}{
\begin{tabular}{cccccc}
\toprule
\multicolumn{2}{c}{\multirow{2}{*}{Distribution Shift}}&\multicolumn{2}{c}{FCOS3D}&\multicolumn{2}{c}{PGD}\\
&&mAP&NDS&mAP&NDS\\
\midrule
\multicolumn{2}{c}{Clean}&29.8&37.7&31.7&39.3\\
\midrule
\multirow{4}{*}{Environment}&Snow&2.07&5.87&3.45&13.21\\
&Rain&16.98&29.08&18.24&29.60\\
&Fog&17.97&29.50&17.48&28.21\\
&Sunlight&27.31&36.92&30.68&38.59\\
\midrule
\multirow{3}{*}{Sensor}&Uniform Noise&5.53&11.69 &2.60&10.75\\
&Gaussian Noise&3.62&6.47&4.57&7.30\\
&Impulse Noise&6.35&10.97&5.86&15.48 \\
\midrule
\multirow{4}{*}{Object}&Scale&6.57&19.53&6.82&21.55 \\
&Shear&21.57&31.63&23.63&33.24\\
&Rotation&13.25&22.52&12.46&20.46\\
&Motion Blur&9.34&20.30&8.19&18.95\\
\bottomrule
\end{tabular}
}}
\label{table6}
\end{table}


\begin{table*}[t]
\small
\caption{The performance of ADs on pre-crash safety-critical scenarios. CR: collision rate, OR: average distance driven out of road, DR: degree of deviation from the pre-defined route, RC: average percentage of route completion.}
\centering
{\resizebox{\linewidth}{!}{
\begin{tabular}{c|c|cccccccc|c}
\toprule
\multirow{3}{*}{Metric}&\multirow{3}{*}{Algo.}&\multicolumn{8}{c}{Pre-crash safety-critical scenarios}&\multirow{3}{*}{Avg.}\\
&&\makecell[c]{Straight\\ Obstacle}&
\makecell[c]{Turning\\ Obstacle}&\makecell[c]{Lane\\
Changing}&\makecell[c]{Vehicle\\
Passing}&\makecell[c]{Red-light\\
Running}&\makecell[c]{Unprotected\\
Left-turn}&\makecell[c]{Right-\\
turn}&\makecell[c]{Dynamic Object\\
Crossing}\\
\midrule
\multirow{3}{*}{CR}&PPO&0.0&0.0&1.0&1.0&0.0&0.0&0.0&0.0&0.25\\
&SAC&1.0&1.0&0.0&0.0&1.0&1.0&0.0&1.0&0.63\\
&TD3&1.0&0.25&0.75&0.0&1.0&1.0& 1.0&1.0&0.75 \\
\midrule
\multirow{3}{*}{OR}&PPO&0.0&0.0&0.0&0.0&0.0&0.0&0.0&0.0&0.00\\
&SAC&0.44&0.71&0.0&0.0&0.0&0.0&0.0&0.0&0.14\\
&TD3&0.0&0.0&0.34&0.0&0.0&0.0&0.0&0.0& 0.04\\
\midrule
\multirow{3}{*}{DR}&PPO&0.43&0.59&0.71&0.44&0.43&0.45&0.51&0.44& 0.50\\
&SAC&0.0&0.77&1.12&1.23&0.41&0.45&0.70&0.44& 0.64\\
&TD3&0.35&0.72&0.44&1.04&0.68&0.33&0.40&0.35&0.54\\
\midrule
\multirow{3}{*}{RC}&PPO&1.0&1.0&0.66&0.60&1.0&1.0&1.0&1.0& 0.91\\
&SAC&0.74&0.77&1.0&1.0&0.53&0.46&1.0&0.62& 0.77\\
&TD3&0.74&0.95&0.75&1.0&0.53&0.47&0.50&0.62& 0.70\\
\bottomrule
\end{tabular}
}}
\label{tablescenario1}
\end{table*}
\section{System-Level Evaluation on Safety-critical Scenarios}
%
Despite extensive experiments confirming that static safety-critical scenarios can lead to errors in intelligent modules, AI component-level errors do not necessarily lead to system level effects. 
For example, the perception model in AD misses a sounding vehicle on the left lane or misrecognizances it as a pedestrian. However, sometimes such perception error does not decisively affect the planing and control of the AD.
In fact, dynamic safety-critic scenarios can often directly affect the planing and control module of AD. 
For instance, a sounding vehicle on the left lane suddenly and maliciously intrudes into the ego vehicle's lane, which can lead to a collision or the ego vehicle braking sharply. 
Therefore, we need to further investigate the safety of the AD system on dynamic safety-critical scenarios.

\subsection{Autonomous Driving Scenarios}
To evaluate the scenario generation methods, we use 2 groups of autonomous driving data obtained from CARLA simulator.
First, we adopt knowledge-based scenario generation method, i.e., pre-defined rules, to obtain pre-crash safety-critical scenarios, which use 8 representative hand-crafted scenarios of pre-crash traffic summarized by the National Highway Traffic Safety Administration (NHTSA): Straight Obstacle, Turning Obstacle, Lane Changing, Vehicle Passing, Red-light Running, Unprotected Left-turn, Right-turn and Crossing Negotiation. 
These 8 scenarios are implemented by SafeBench~\cite{xu2022safebench}.
For each scenario, it designs 4 diverse driving routes that vary in terms of road layouts, surrounding environments, etc. Such manually designed functional test item poses significant challenges to ADs.
Second, we also use natural driving datasets Town02 and Town05, which merge ``Scenario6'' and ``Scenario7'' in SafeBench, spanning across CARLA maps Town02 and Town05.
Note that these two datasets Town02 and Town05 differ from Town05 and Town06 used in the adversarial attack evaluation in Sec.\ref{sec3}.
In Town02 and Town05, a total of 15 routes are predefined, and the AV navigates these predefined routes through different junctions to form a total of 150 scenarios, 70 scenarios for Town02, 80 scenarios for Town05.
In the experiments, we employ driving policy-based adversarial generation methods to transform natural driving scenarios into accident-prone safety-critical scenarios, aiming to evaluate the safety of ADs.
\subsection{Baselines}
For autonomous driving scenarios, the baselines includes: (1) \textbf{Autopilot} method utilizes a rule-based autopilot policy implemented in the Carla Simulator to generate realistic traffic flows;
(2) We select 3 representative deep RL methods for evaluation, including a stochastic off-policy method Soft Actor-Critic (\textbf{SAC}), a stochastic on-policy algorithm Proximal Policy Optimization (\textbf{PPO}) and a deterministic off-policy approach Twin Delayed DDPG (\textbf{TD3}).
(3) We use driving policy-based adversarial generation method to optimize and generate safety-critical scenarios.
Specifically, \textbf{Adv.PPO} method employs an adversarial surrounding vehicles based on PPO reinforcement learning agent that aims to reach a potential conflict point with the ego vehicle by utilizing the adversarial reward function as Eq.\ref{eq6}.

\subsection{Evaluation metrics}
We follow existing work~\cite{li2022metadrive} and CARLA leaderboard, and consider 4 evaluation metrics, collision rate (CR), average distance driven out of road (OR), distance to route (DR) and average percentage of route completion (RC), focusing on serious violations of traffic rules and functional ability of AD.
Specifically, CR is the percentage of routes in which the agent collided while traversing an intersection.
OR measures how far the ego vehicle strays from the road during driving of ego vehicle.
DR represents the degree of deviation from the pre-defined route during the driving of ego vehicle.
RC is the percentage of the route completed by an agent before it gets blocked or deviates from the route. 
RC can be utilized to detect whether the vehicle is in motion. ‌If RC is not triggered‌, it indicates that the vehicle is either in a stationary state or stuck, even though other performance metrics remain satisfactory.
\begin{table}[t]
\small
\caption{Performance of the ADs against scenario generation algorithm Adv.PPO on Town02 and Town05 maps.}
\centering
{\resizebox{\linewidth}{!}{
\begin{tabular}{c|c|ccc|ccc}
\toprule
\multirow{2}{*}{AV}&\multirow{2}{*}{Surr. AV}&\multicolumn{3}{c}{Town02}&\multicolumn{3}{c}{Town05}\\
&&CR&OR&DR&CR&OR&DR\\
\midrule
Autopilot&Autopilot&0.01&0.00&0.03&0.00&0.00&0.07\\
PPO&Autopilot&0.34&0.04&0.19&0.11&0.01&0.16\\
Autopilot&Adv. PPO&0.17&0.00&0.07&0.21&0.00&0.06\\
PPO&Adv. PPO&0.81&0.05&0.22&0.59&0.03&0.12\\
\bottomrule
\end{tabular}
}}
\label{tablescenario2}
\end{table}
\subsection{Experimental Setting}
We evaluate the performance of the ADs on the pre-crash safety-critical scenarios, Town02 and Town05, respectively, by using safety-critical scenario generation methods.
For the pre-crash safety-critical scenarios, Xu et al.,~\cite{xu2022safebench} design a trigger-based scenario generation.
That is, the behaviors of the surrounding objects in the scenario such as vehicles or pedestrians are manually designed in advance.
The ADs trigger the pre-defined behaviors when the distance between the surrounding object and the ego vehicle is less than a threshold.
Unlike the trigger-based scenarios in the pre-crash safety-critical scenarios, Adv.PPO relies on random traffic flows and controls critical surrounding vehicles in the scenario, thereby prompting the adversarial vehicle to attack the ego vehicle and create safety-critical scenarios.
Thus, Adv.PPO transforms the generation of safety-critical scenarios into a planning and control problem of ADs, encouraging them to execute reasonable and flexible attack behaviors.
\begin{figure*}[ht]
\centering     
\subfigure[Adversarial Cut In]{\label{fig3:0}\includegraphics[width=0.32\linewidth]{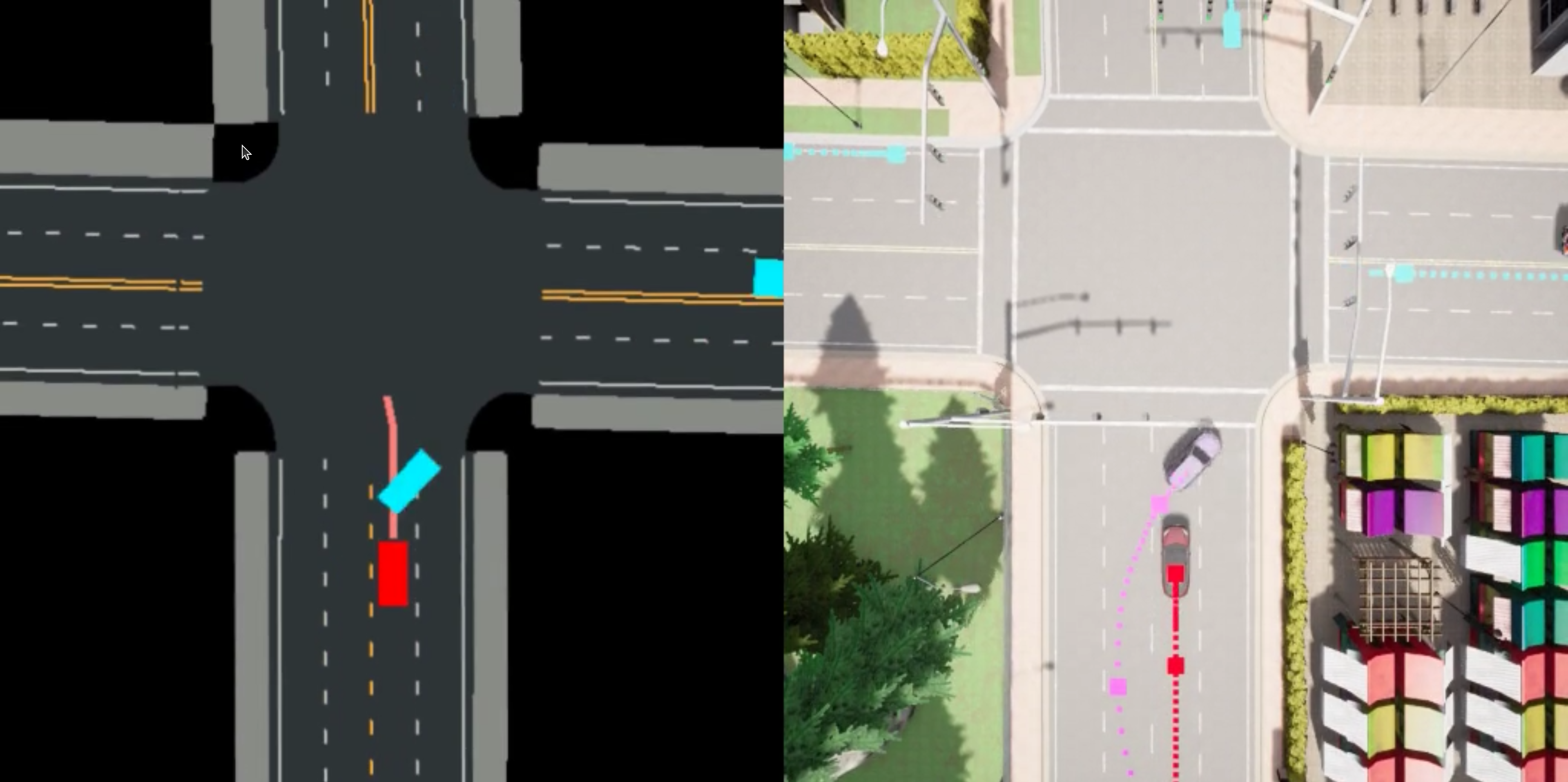}}
\subfigure[Adversarial Left-turn]{\label{fig3:0}\includegraphics[width=0.32\linewidth]{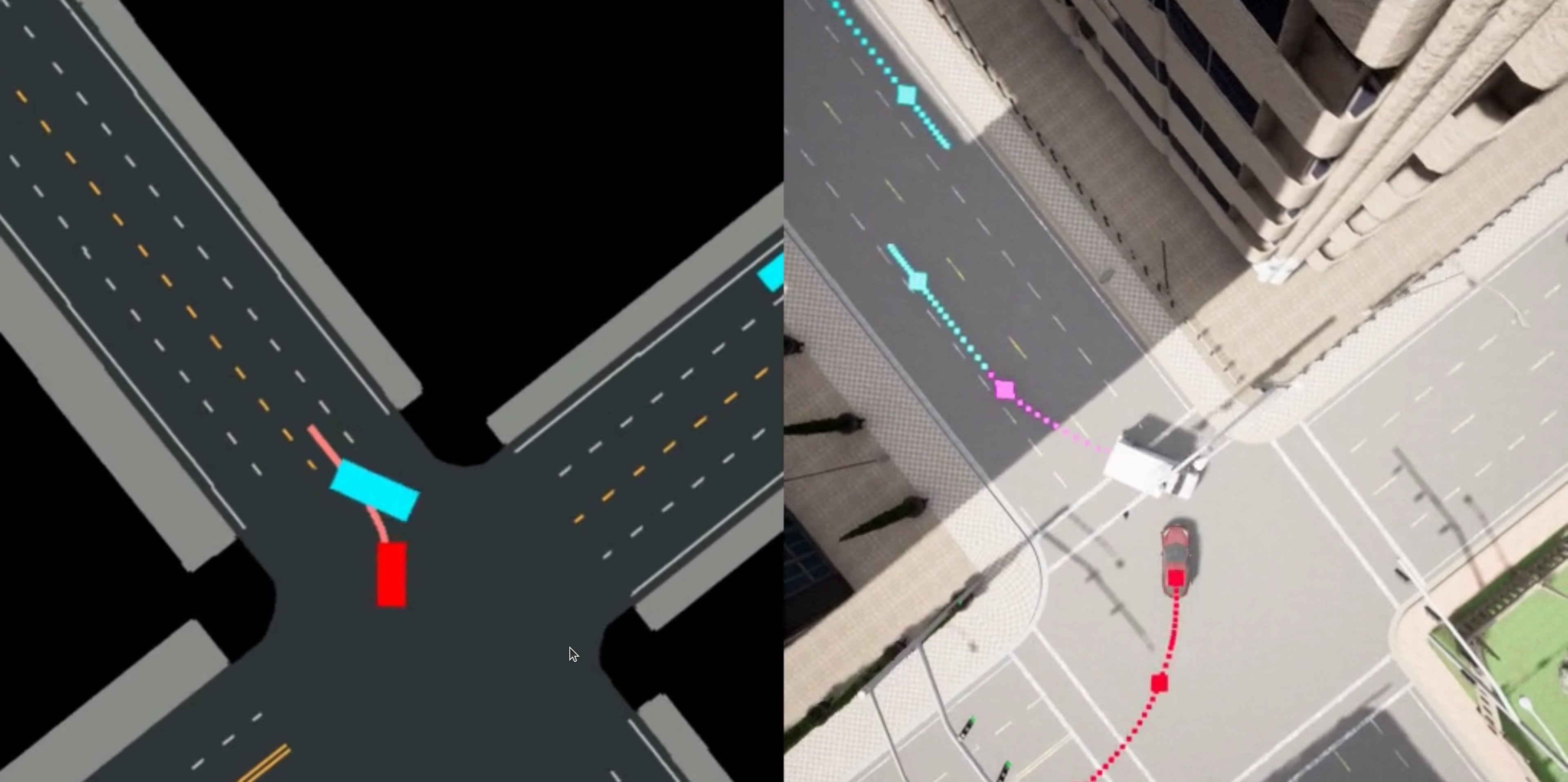}}
\subfigure[Adversarial Right-turn]{\label{fig3:1}\includegraphics[width=0.32\linewidth]{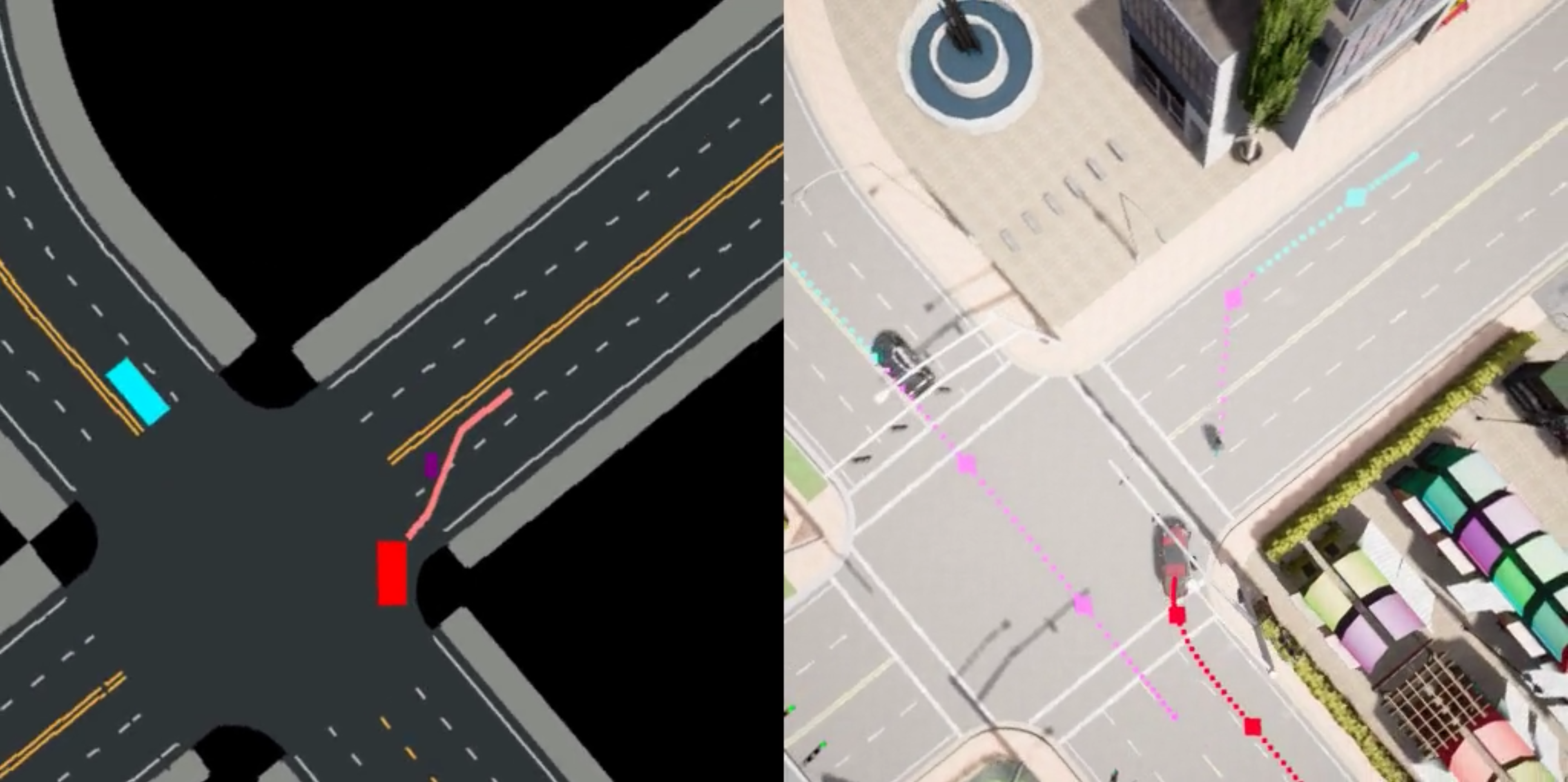}}
\caption{Visualization of the performance of the Autopilot under the scenario generation algorithm Adv.PPO on Twon02. In the bird’s eye view, the red box is the ego vehicle controlled by Autopilot and the adversarial surrounding vehicle is controlled by Adv.PPO with blue colors.}
\label{scs1}
\end{figure*}
\begin{figure*}[ht]
\centering     
\subfigure[Adversarial Opposite Vehicle]{\label{fig3:0}\includegraphics[width=0.32\linewidth]{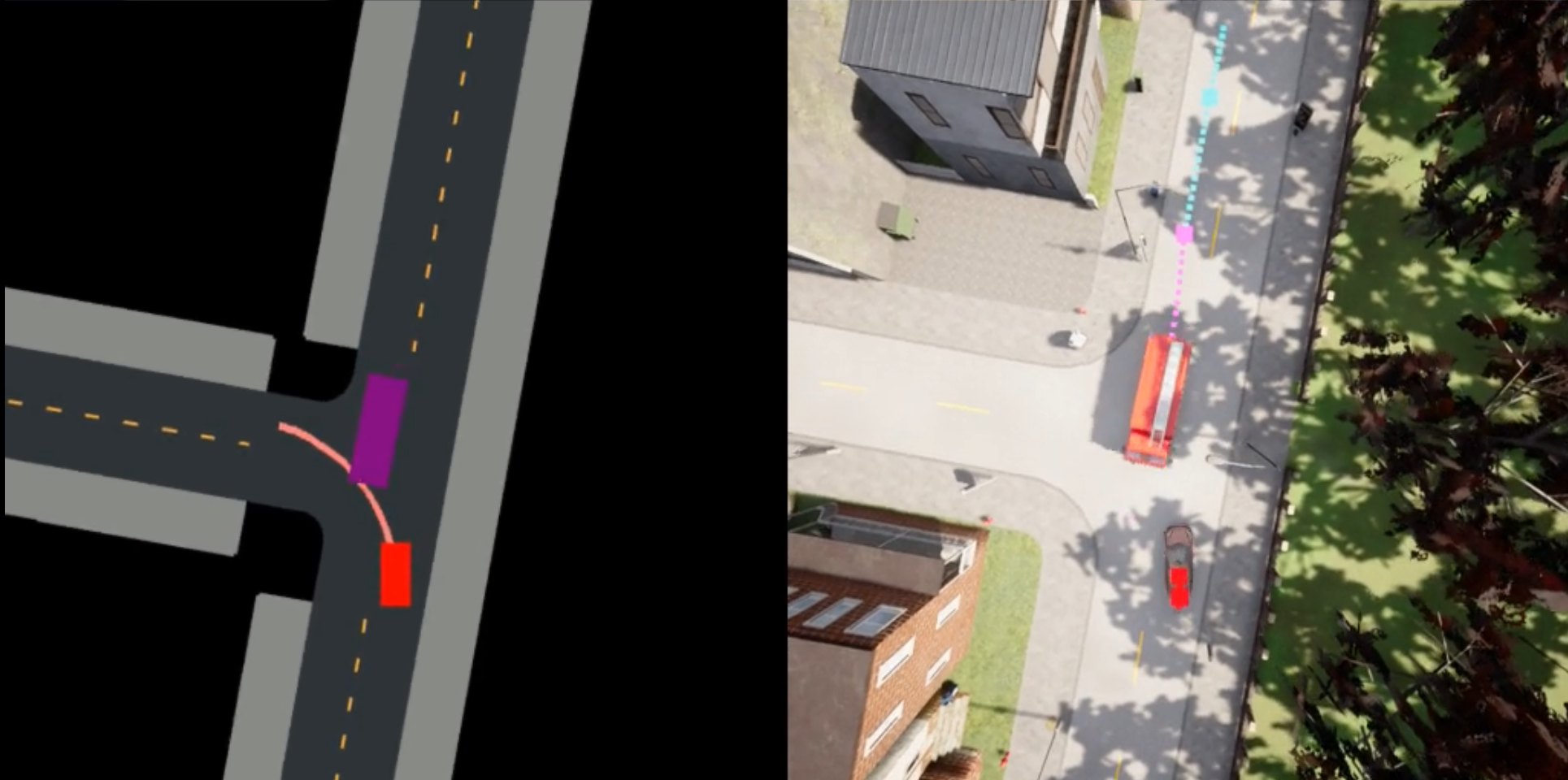}}
\subfigure[Adversarial Right-turn]{\label{fig3:0}\includegraphics[width=0.32\linewidth]{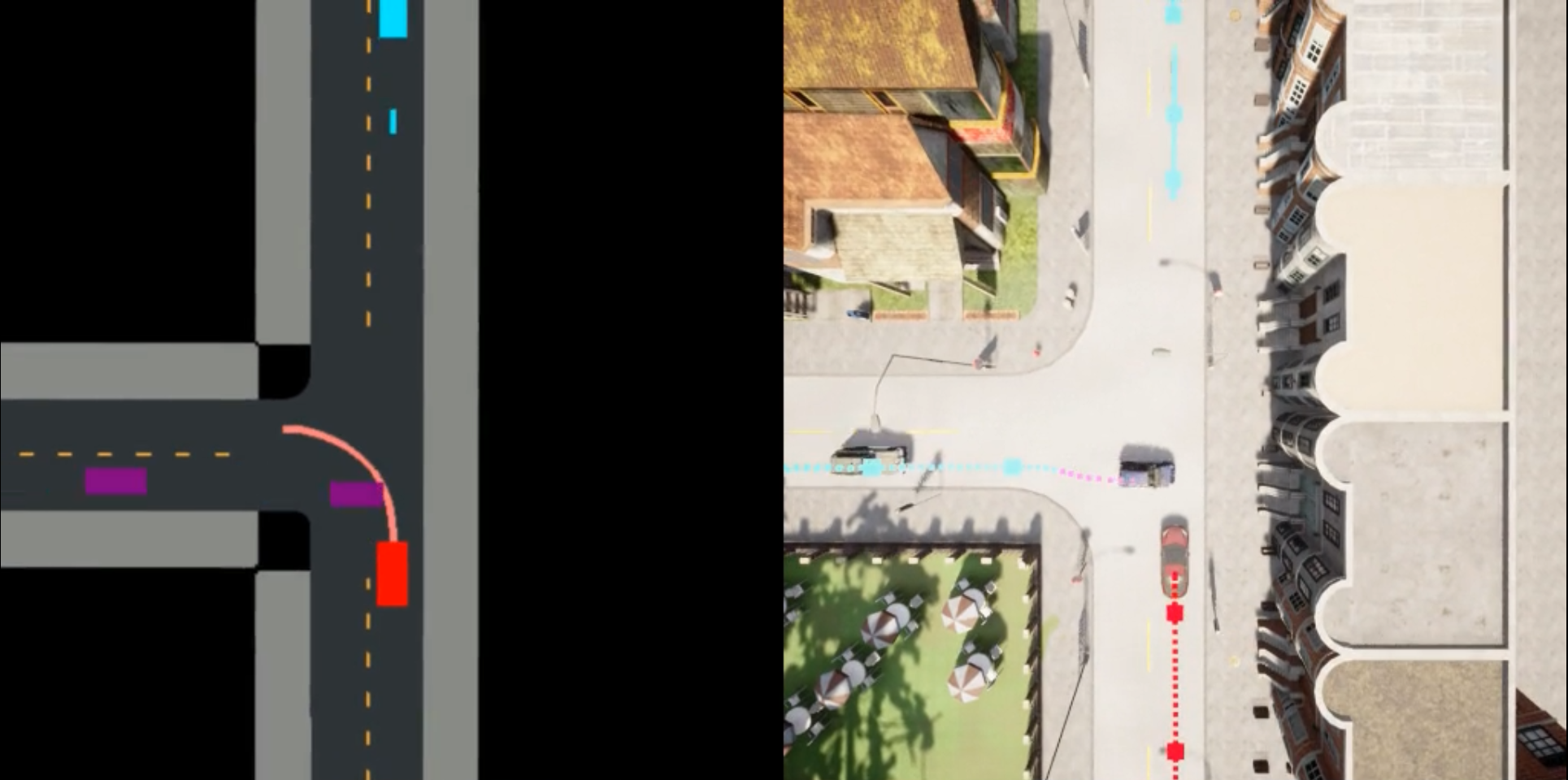}}
\subfigure[Adversarial Pedestrian Crossing]{\label{fig3:1}\includegraphics[width=0.32\linewidth]{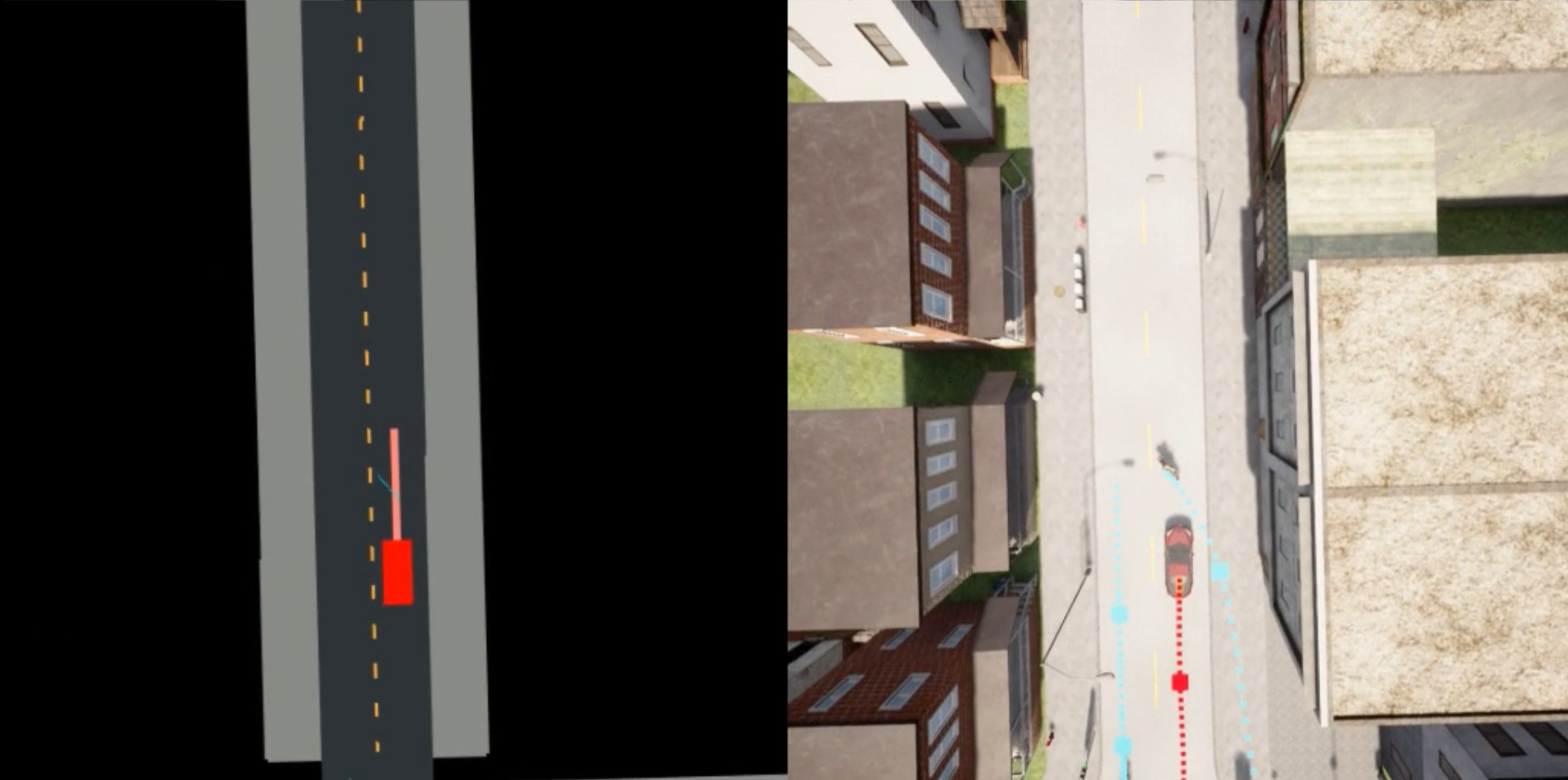}}
\caption{Visualization of the performance of the Autopilot under the scenario generation algorithm Adv.PPO on Twon05. The red box is the
ego vehicle and the adversarial surrounding
vehicles are in brown color.}
\label{scs2}
\end{figure*}
\subsection{Experimental Results}
Tab.\ref{tablescenario1} reports the performance of some ADs on safety-critical scenarios.
The experimental results reveal that these pre-trained reinforcement learning algorithms exhibit functional safety vulnerabilities when tested on manually designed rule-based safety-critical scenarios. For instance, the TD3 algorithm demonstrates an average collision rate of 75\%‌. Although the PPO algorithm shows relatively better performance, collisions still occur in pre-defined Lane Changing and Vehicle Passing scenarios.


Unlike Tab.\ref{tablescenario1} which evaluates trigger-based safety-critical scenarios, Tab.\ref{tablescenario2} reports the performance of ADs for the adversarial-based scenario generation method on Town02 and Town05.
For the ego vehicle, we use Autopilot and the RL method PPO, respectively.
For the surrounding traffic participants, we use Adv.PPO to transform natural driving scenarios into accident-prone scenario to evaluate the safety of ADs.
As shown in Tab.\ref{tablescenario2}, Autopilot demonstrates near-collision-free performance under non-adversarial conditions, while the PPO algorithm also maintains adequate safety levels in most scenarios.
When deploying Adv.PPO to generate accident-prone scenario, Autopilot exhibits frequent collision events, while the PPO algorithm demonstrates collision rates of 81\% and 56\% on Town02 and Town05 respectively.

It can be seen that the safety-critical scenario generation methods can directly and effectively threaten the ADs, while different ADs have different feedbacks for the different safety-critical scenarios. 
Therefore, constructing safety-critical scenarios is of great significance for studying the safety of ADs.

\subsection{Analysis of Safety-Critical Scenarios}
In Figs.\ref{scs1} and \ref{scs2}, we visualize scenarios generated by Adv.PPO  method in different scenarios on Town02 and Town05, respectively.
In these scenarios, the ego vehicle is controlled by Autopilot, and sounding vehicle is planned by Adv.PPO to produce attack behavior for the ego vehicle.
In the bird's eye (BEV) view, the red box is the ego vehicle and the adversarial surrounding vehicles are in blue and brown colors.
For example, as shown in Fig.\ref{scs1}(a), the Adv.PPO controls the surrounding vehicles to perform a malicious cut-in behavior toward the ego vehicle; in Fig.\ref{scs1}(c), the ego vehicle encounters a pedestrian suddenly crossing the road controlled by Adv.PPO as the ego vehicle making a right turn.
These safety-critical scenarios are relatively rare and difficult to collect in conventional driving scenarios. Through studying the performance of the ADs in these safety-critical scenarios, its safety performance can be effectively evaluated.

\section{Conclusions and Future Work}
In this work, we conduct a comprehensive investigation of current research on autonomous driving scenarios and argue that the safety-critical scenarios as corner cases pose significant threats to the safety and robustness of autonomous driving. 
We present a categorization of safety-critical scenario  which includes static traffic scenarios and dynamic traffic scenarios. Specifically, static traffic scenarios encompass adversarial attack scenarios and natural distribution shifts. Dynamic traffic scenarios, on the other hand, primarily involve the use of scenario generation algorithms to create accident-prone scenarios.
To bridge the AI-to-system semantic gap, we assess the safety and robustness of perception modules, particularly object detection task, under various adversarial attack methods and distribution shift scenarios.
Furthermore, we evaluate the safety of autonomous driving system for accident-prone scenarios generated by scenario generation algorithms. 
Our research reveals safety and robustness challenges in autonomous driving systems, spanning from perception modules to system-level under safety-critical scenarios.
Despite the progress made in this study, there are still several limitations and avenues for future research:
\begin{itemize}

\item \textbf{‌Robustness‌ of multi-modal perception.} We only focus on the camera-based object detection task.
However, LiDAR-based detection and multi-modal fusion approaches~\cite{bai2022transfusion,xie2023robobev} deserve further investigation to understand their robustness in safety-critical scenarios.
In addition, designing robust perception algorithms that can handle safety-critical scenarios is crucial for ensuring the safety and robustness of autonomous driving systems.

\item \textbf{‌Enhancing robustness through safety-critical scenario.} The safety-critical scenarios expose vulnerabilities in autonomous driving. Therefore, leveraging safety-critical scenario data as training inputs can effectively enhance the robustness of autonomous driving systems~\cite{hanselmann2022king}.
These generated scenarios can be utilized to train autonomous driving systems, thereby enhancing their safety. However, the accident-prone scenarios are difficult to collect in the real world. Accident-prone driving scenario algorithms~\cite{zhang2023cat} provide a promising solution to produce large-scale accident-prone scenarios in simulation environments or by utilizing real-world driving data.
%

\item \textbf{‌Controllable and diverse scenario generation.} The accident-prone scenario generation methods often produce uncontrollable and insufficiently diverse scenarios. 
Developing more sophisticated algorithms that can generate controllable and diverse safety-critical scenarios will be essential for a more comprehensive evaluation of autonomous driving systems.
World model-based scenario generation algorithms~\cite{wei2024editable,wen2024panacea,yang2024drivearena} have shown potential in generating diversity and photo-realistic corner cases and could be explored further.

\item \textbf{‌The safety and robustness of advanced end-to-end and VLM autonomous driving.} The evaluations of autonomous driving systems in safety-critical scenarios are based on reinforcement learning algorithms.
Recent advancements in end-to-end learning~\cite{wu2022trajectory,jiang2023vad} and vision-language models~\cite{shao2024lmdrive,ma2024dolphins} have demonstrated superior driving performance. The safety and robustness of these advanced autonomous driving systems on safety-critical scenarios need to be further explored.
\end{itemize}

\ifCLASSOPTIONcaptionsoff
  \newpage
\fi



\bibliographystyle{IEEEtran}
\bibliography{bibtex/bib/IEEEexample}
\begin{IEEEbiography}
[{\includegraphics[width=1in,height=1.25in]{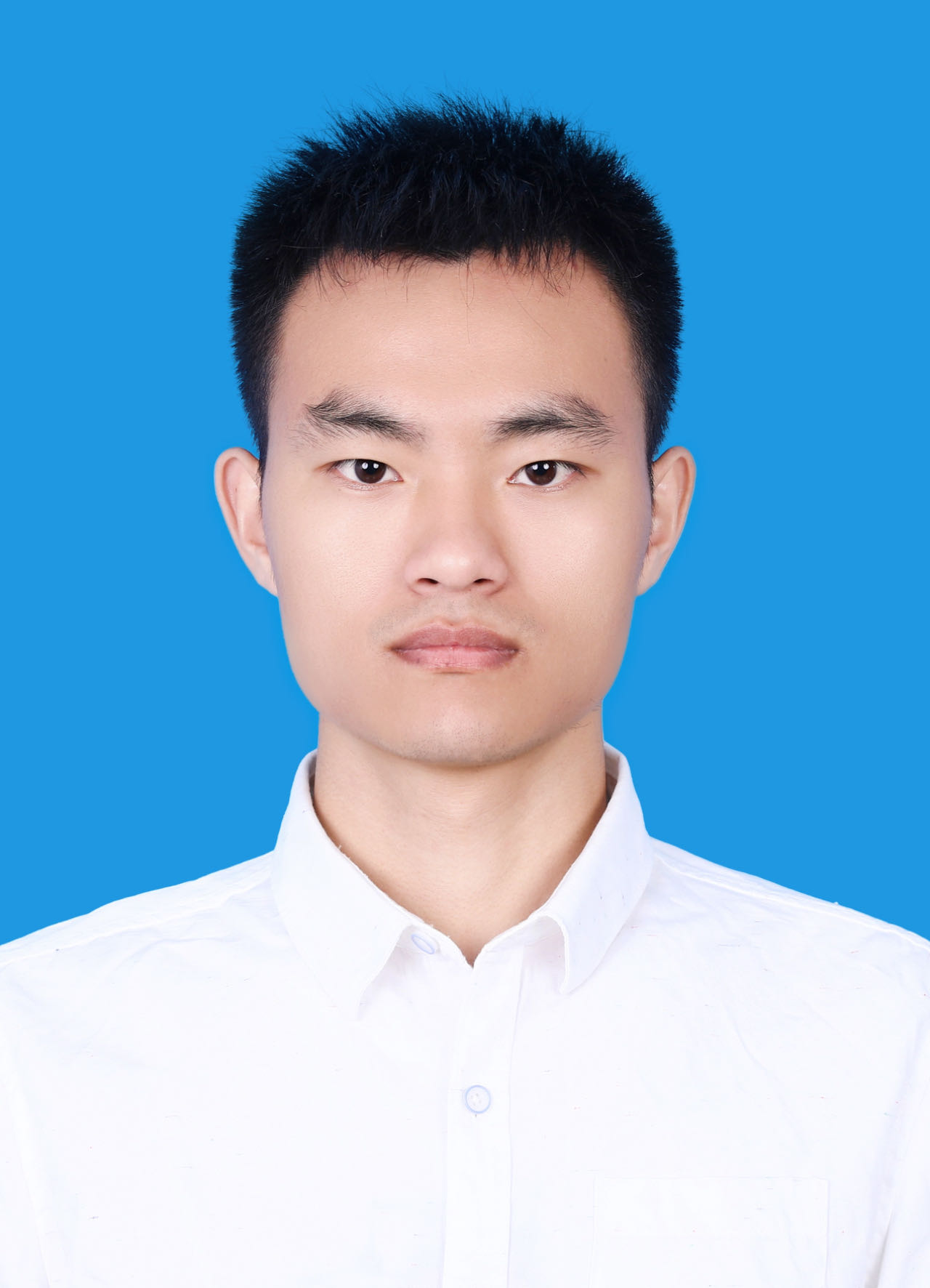}}]
{Jingzheng Li} received his Ph.D degree in 2024 from School of Computer Science and Engineering, Beihang University. He is currently a Research Scientist
with Zhongguancun Laboratory, Beijing, China.
His research focus on robust deep learning under data distribution shift, e.g., domain adaptation, out-of-distribution generalization. Recently, his interest is the safety and robustness of autonomous vehicles in complex scenarios.
\end{IEEEbiography}
 
\begin{IEEEbiography}
[{\includegraphics[width=1in,height=1.25in]{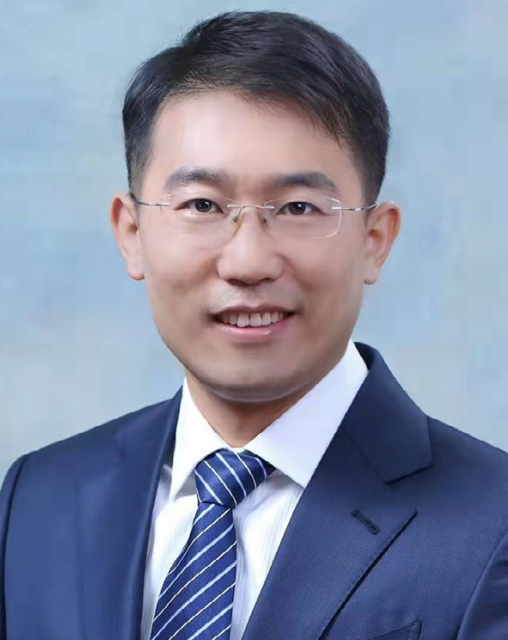}}]
{Xianglong Liu} received the B.S. and Ph.D. degrees, under the supervision of Prof. Wei Li. He visited the DVMM Laboratory, Columbia
University, as a joint Ph.D. Student, supervised by Prof. Shih-Fu Chang. He is currently a Full Professor with the School of Computer Science and Engineering, Beihang University. His research interests include fast visual computing (e.g., large-scale search/understanding) and robust deep learning (e.g., network quantization, adversarial attack/defense, and few-shot learning). He received the NSFC Excellent Young Scientists Fund and was selected for the 2019 Beijing Nova Program, the MSRA StarTrack Program, and the 2015 CCF Young Talents Development Program.
\end{IEEEbiography}

\begin{IEEEbiography}
[{\includegraphics[width=1in,height=1.25in]{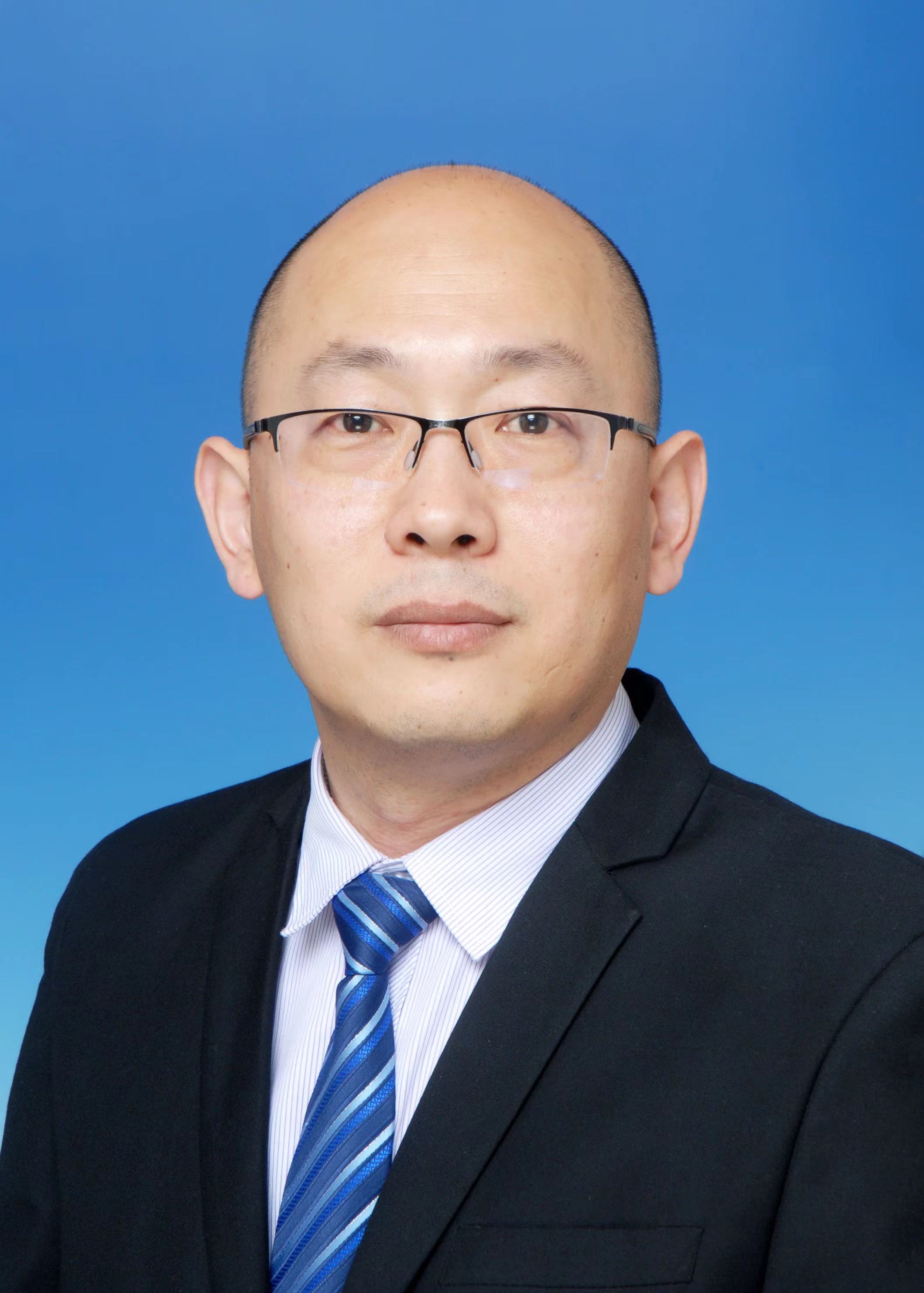}}]
{Shikui Wei} received the B.E. degree from Hebei University in 2003 and the Ph.D. degree in signal and information processing from Beijing Jiaotong University (BJTU), China, in 2010. From 2010 to 2011, he worked as a Research Fellow with the School of Computer Science and Engineering, Nanyang Technological
University, Singapore. He is currently a Full Professor with the Institute of Information Science, BJTU. His research interests include computer vision, image/video analysis and retrieval. He is a senior member of the IEEE.
\end{IEEEbiography}

\begin{IEEEbiography}
[{\includegraphics[width=1in,height=1.25in]{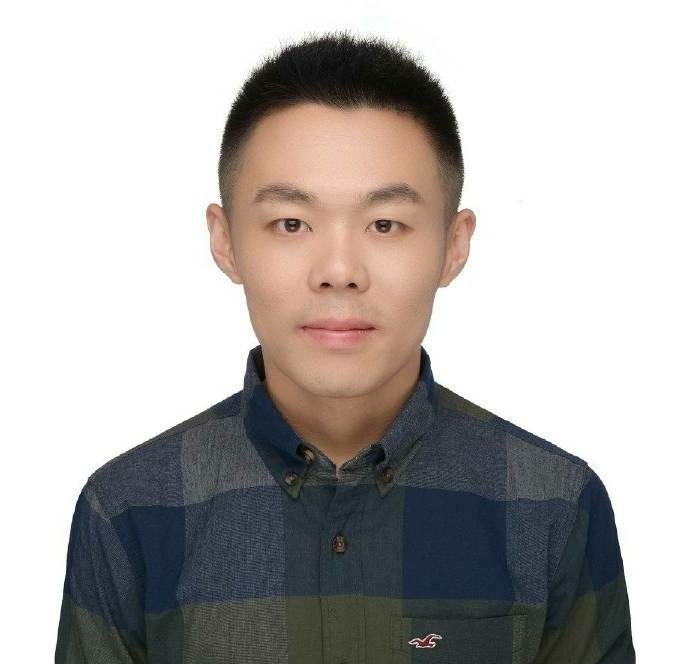}}]
{Zhijun Chen}
received his Ph.D. in 2024 from the School of Computer Science, Beihang University. His primary research interests include ensemble algorithms for large language models, weak supervision learning, and its security.
\end{IEEEbiography}

\begin{IEEEbiography}
[{\includegraphics[width=1in,height=1.25in]{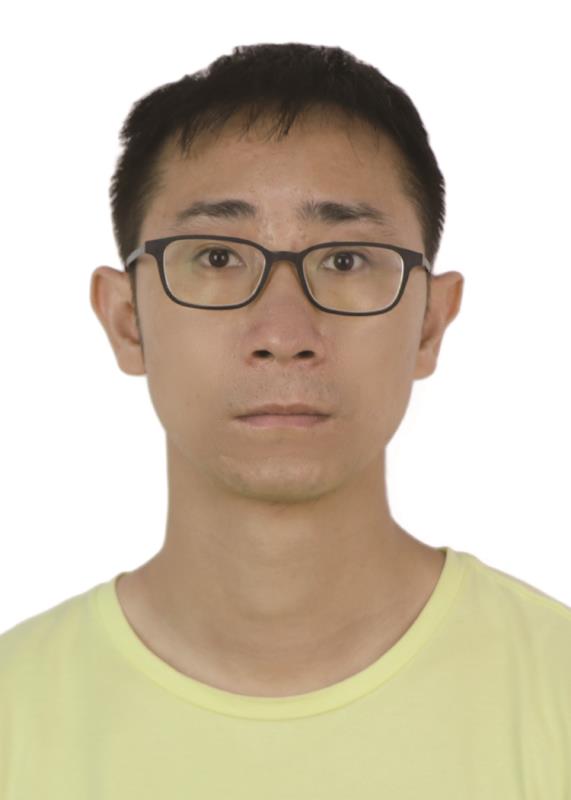}}]
{Bing Li}
received his Ph.D. degree in 2024 from the School of Computer Science, Beihang University. He is currently a Postdoctoral Researcher at A*STAR, Singapore. His research focuses on video understanding (e.g., video classification, object tracking) and the safety of 3D detection in autonomous driving.
\end{IEEEbiography}

\begin{IEEEbiography}
[{\includegraphics[width=1in,height=1.25in]{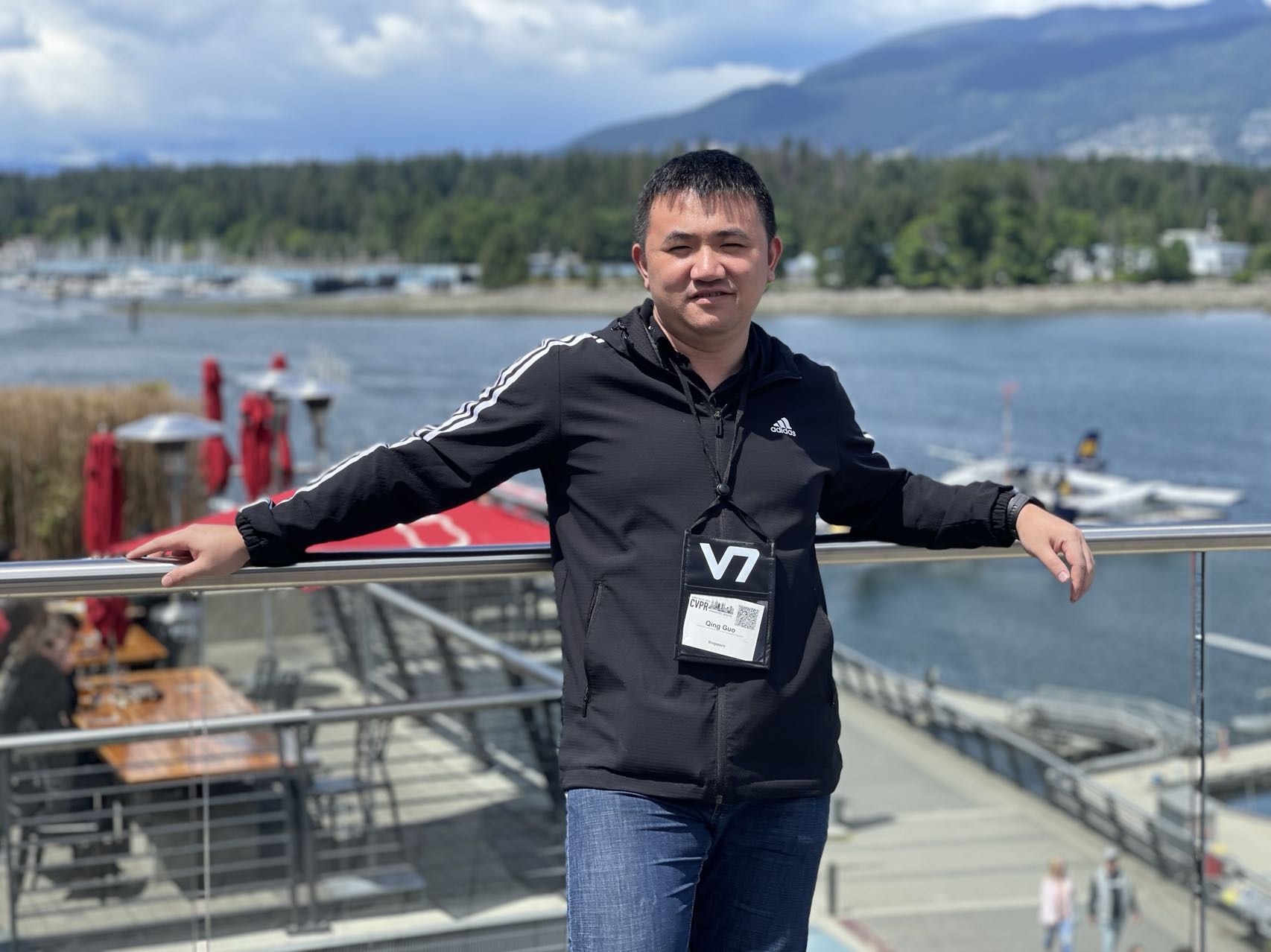}}]
{Qing Guo}
is currently a senior research scientist and principal investigator (PI) at the Center for Frontier AI Research (CFAR), A*STAR in Singapore. He is also an adjunct assistant professor at the National University of Singapore (NUS).  His research mainly focuses on computer vision, AI security, adversarial attack, and robustness.
\end{IEEEbiography}

\begin{IEEEbiography}
[{\includegraphics[width=1in,height=1.25in]{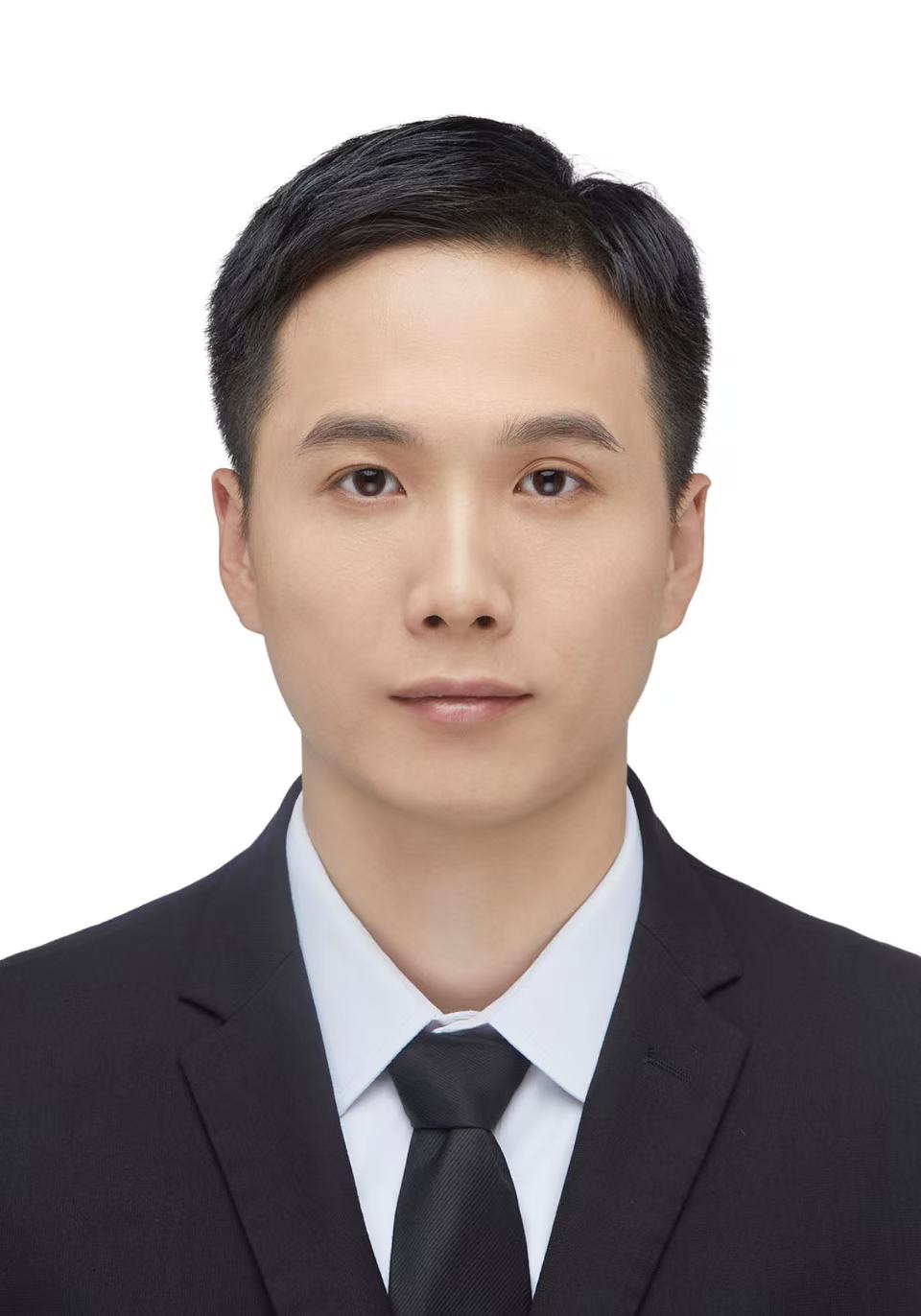}}]
{Xianqi Yang}
is now an engineer in Zhongguancun Laboratory, Beijing, China. He received the master degree in 2023 from Beihang University, supervised by Prof. Qing Gao and Prof. Kexin Liu. His research interests are blockchain, autonomous driving security and industrial internet security.
\end{IEEEbiography}

\begin{IEEEbiography}
[{\includegraphics[width=1in,height=1.25in]{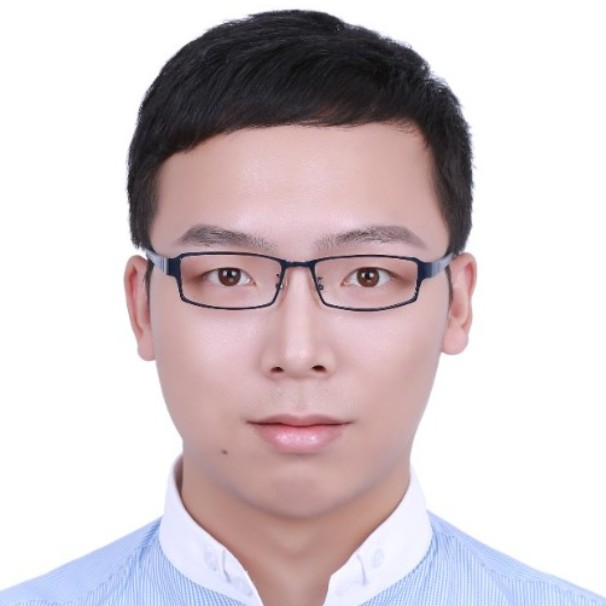}}]
{Yanjun Pu}
received his Bachelor's degree in Computer Science \& Engineering from Beihang University in 2016 and earned his Ph.D. in Software Engineering from the same institution in 2023. He currently serves as an assistant researcher at Zhongguancun Laboratory. His research interests include Educational Data Mining, Service Computing, and Autonomous Driving Technologies.
\end{IEEEbiography}

\begin{IEEEbiography}
[{\includegraphics[width=1in,height=1.25in]{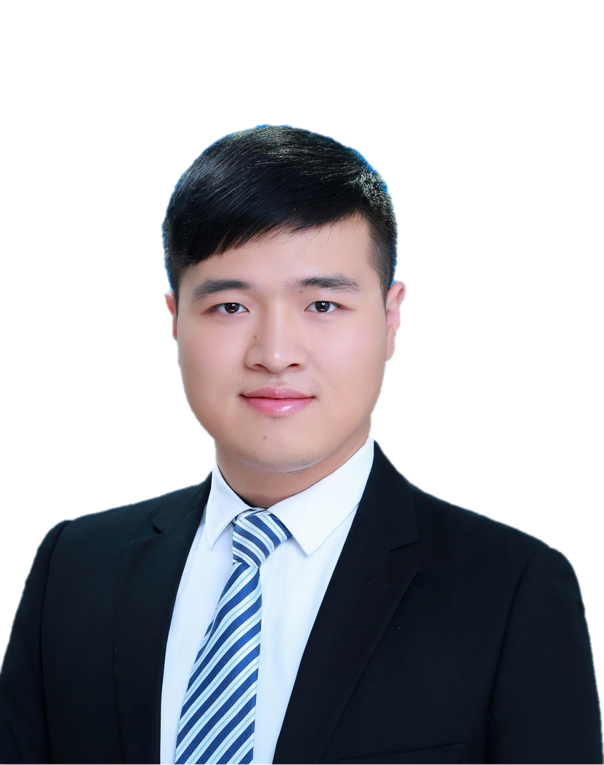}}]
{Jiakai Wang}
is now a Research Scientist in Zhongguancun Laboratory, Beijing, China. He received the Ph.D. degree in 2022 from Beihang University (Summa Cum Laude), supervised by Prof. Wei Li and Prof. Xianglong Liu. Before that, he obtained his BSc degree in 2018 from Beihang University (Summa Cum Laude). His research interests are Trustworthy AI in  computer Vision (mainly) and Multimodal Machine Learning, including Physical Adversarial Attacks and Defense and Security of Practical AI.
\end{IEEEbiography}

\end{document}